%% file: paper.tex
\documentclass[10pt,twocolumn,letterpaper]{article}

\usepackage{cvpr}
\usepackage{times}
\usepackage{epsfig}
\usepackage{graphicx}
\usepackage{amsmath}
\usepackage{amssymb}
\usepackage{symbols}
\usepackage{mathbbol}
\usepackage{svg}
\usepackage{microtype}
\usepackage[ruled,linesnumbered]{algorithm2e}
\usepackage[font=small, labelfont=bf]{caption}
\usepackage{multirow}

\usepackage{algorithmic}
\usepackage{bbm}
\usepackage{mathbbol}
\usepackage{booktabs} 
\usepackage{setspace}

\usepackage[pagebackref=true,breaklinks=true,letterpaper=true,colorlinks,bookmarks=false]{hyperref}

\cvprfinalcopy 

\def\cvprPaperID{9158} 

\ifcvprfinal\fi
\begin{document}

\title{Can We Learn Heuristics For Graphical Model Inference Using Reinforcement Learning?}

\author{Safa Messaoud, Maghav Kumar, Alexander G. Schwing\\
University of Illinois at Urbana-Champaign\\
{\tt\small \{messaou2, mkumar10, aschwing\}@illinois.edu}}

\maketitle

\input{abs}
\input{intro}

\input{rel}

\input{app}

\input{exp}

\input{conc}

{\small
\bibliographystyle{ieee_fullname}
\bibliography{alex}
}
\clearpage
\onecolumn
\input{supplementary}


\end{document}

%% file: abs.tex
\begin{abstract}
Combinatorial optimization is frequently used in computer vision. For instance, in applications like semantic segmentation, human pose estimation and action recognition, programs are formulated for 
solving inference in Conditional Random Fields (CRFs) to produce a structured output that is consistent with visual features of the image. However, solving inference in CRFs is in general intractable,  and approximation methods are computationally demanding and limited to unary, pairwise and hand-crafted forms of higher order potentials. In this paper, we show that we can learn program heuristics, \ie, policies, for solving inference in higher order CRFs for the task of semantic segmentation, using reinforcement learning. Our method solves inference tasks efficiently without imposing any constraints on the form of the potentials. We show compelling results on the Pascal VOC and  MOTS datasets.
\end{abstract}

%% file: intro.tex
\vspace{-0.5cm}
\section{Introduction}
\label{sec:intro}

Graphical model inference 
is an important combinatorial optimization task for  robotics and autonomous systems. Despite significant progress in recent years due to increasingly accurate deep net models, challenges such as inconsistent bounding box detection, segmentation or  image classification remain. Those inconsistencies can be  addressed with Conditional Random Fields (CRFs), albeit requiring to solve an  inference task which is of combinatorial complexity. 

%

Classical algorithms to address combinatorial problems come in three paradigms: exact, approximate and heuristic. Exact algorithms are often based on solving an Integer Linear Program (ILP) using a combination of a Linear Programming (LP) relaxation and a  branch-and-bound framework. Particularly for large problems, repeated solving of linear programs is computationally expensive and therefore prohibitive. Approximation algorithms address this concern, however, often at the expense of weak optimality guarantees. Moreover, approximation algorithms often involve manual construction for each problem. 
Seemingly easier to develop are heuristics which are generally computationally fast but  guarantees are hardly provided. In addition, tuning of hyper-parameters for a particular problem instance may be required. A fourth paradigm has been considered since the early 2000s and gained popularity again recently~\cite{ZhangJAIR2000,BoyanJMLR2000,VinyalsNIPS2015,BelloARXIV2016,GuICLR2017,DaiNIPS2017}: learned algorithms. This 
fourth paradigm is based on the intuition that data  governs the properties of the combinatorial algorithm. For instance,  semantic image segmentation always deals with similarly sized problem structures or semantic patterns. It is therefore conceivable that learning to solve the problem on a given dataset uncovers strategies which are close to optimal but hard to find manually, since it is much more effective for a learning algorithm to sift through large amounts of sample problems. To achieve this, in a series of work,  
reinforcement learning techniques were developed~\cite{ZhangJAIR2000,BoyanJMLR2000,VinyalsNIPS2015,BelloARXIV2016,GuICLR2017,DaiNIPS2017} and shown to perform well on a variety of combinatorial tasks from the traveling salesman problem and the knapsack formulation to maximum cut and minimum vertex cover. 

\input{overview_rl}


While the aforementioned learning based techniques have been shown to perform extremely well on classical benchmarks, we are not aware of  results  for inference algorithms in CRFs for semantic segmentation. 
We hence wonder whether we can learn heuristics to address graphical model inference in semantic segmentation problems? 
To study this we develop a new framework for higher order CRF inference for the task of semantic segmentation using a Markov Decision Process (MDP). To solve the MDP, we assess two  reinforcement learning algorithms: a Deep Q-Net (DQN)~\cite{mnih2015human} and a deep net guided Monte Carlo Tree Search (MCTS)~\cite{silver2016mastering}. 

The proposed approach has two main advantages: (1) Unlike traditional approaches, it does not impose any constraints on the form of the CRF terms to facilitate effective inference. We demonstrate our claim by designing detection based higher order potentials that result in computationally intractable classical inference  approaches. (2) Our method is more efficient than traditional approaches as inference complexity is linear in arbitrary potential orders while classical methods have  exponential dependence on the largest clique size in general. This is due to the fact that semantic segmentation is reduced to sequentially inferring the labels of every variable based on a learned policy, without use of any iterative or search procedure. 




We evaluate the proposed approach on two benchmarks: (1) the Pascal VOC semantic segmentation dataset~\cite{EveringhamIJCV2010}, and (2) the MOTS multi-object tracking and segmentation dataset~\cite{voigtlaender2019mots}. We demonstrate that our method outperforms traditional inference algorithms while being more efficient.

%% file: overview_rl.tex
\begin{figure*}[t]
\vspace{-0.5cm}
\centering
\includegraphics[width=0.8\linewidth]{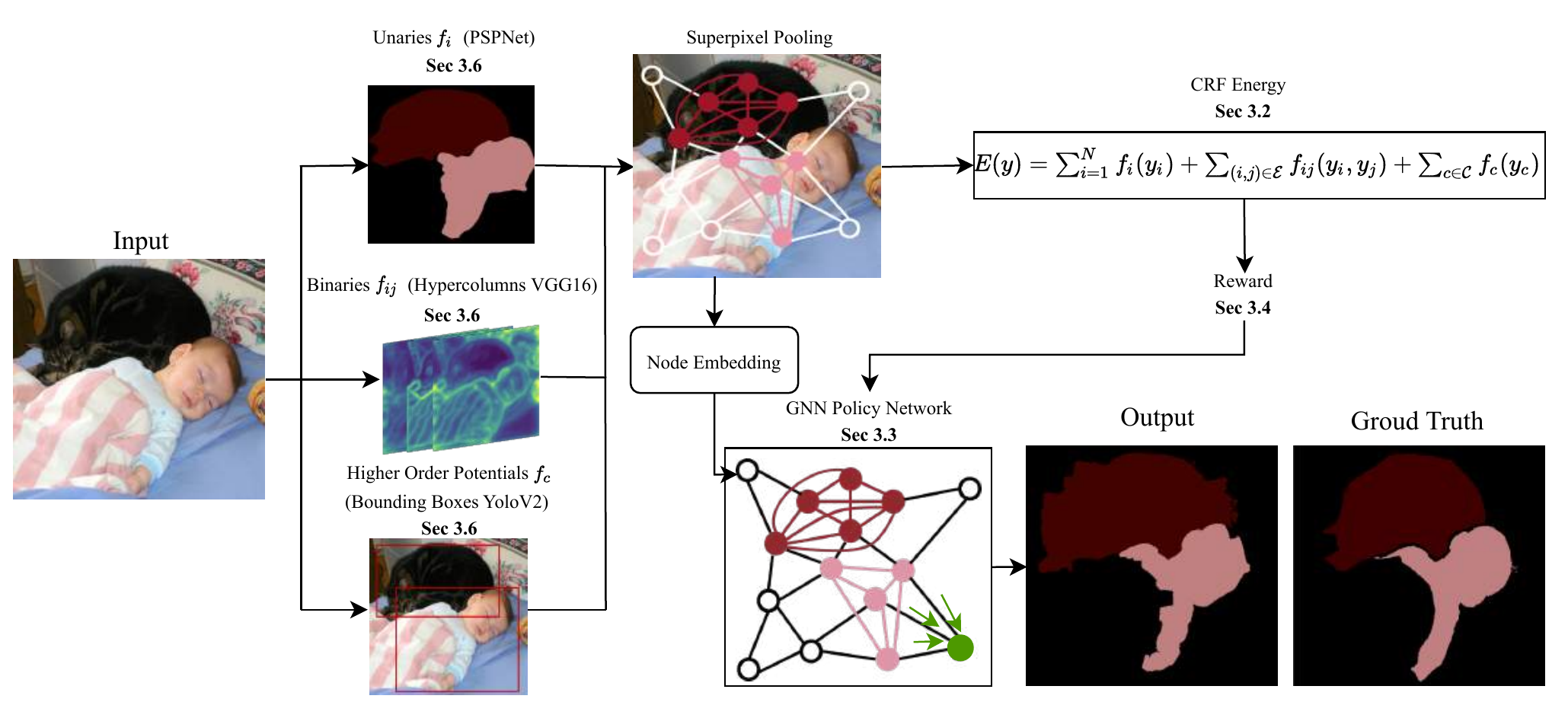}
\caption{Pipeline of the proposed approach. Inference in a higher order CRF is solved using reinforcement learning for the task of semantic segmentation. For Pascal VOC, unaries are obtained from PSPNet~\cite{zhao2017pyramid}, pairwise potentials are computed using hypercolumns from VGG16~\cite{hariharan2015hypercolumns} and higher order potentials are based on detection bounding boxes from YoloV2~\cite{redmon2016you}. The policy network is modeled as a graph embedding network \cite{dai2016discriminative} following the CRF graph structure. It sequentially produces the labeling of every node (superpixel). }
\label{fig:Overview}
\vspace{-0.3cm}
\end{figure*}

%% file: rel.tex
\section{Related Work}
\label{sec:rel}

We first review work on semantic segmentation before discussing learning of combinatorial optimizers. 

\noindent\textbf{Semantic Segmentation:} 
In early 2000, classifiers were locally applied to images to generate segmentations~\cite{KonishiCVPR2000} which resulted in a noisy output. To address this concern, as early as 2004, He \etal~\cite{HeCVPR2004} applied Conditional Random Fields (CRFs)~\cite{Lafferty2001} and multi-layer perceptron features. For inference, Gibbs sampling was used, since MAP inference is NP-hard due to the combinatorial nature of the program. Progress in combinatorial optimization for flow-based problems in the 1990s and early 2000s~\cite{Ford1962,Goldberg1988,Greig1989,Boykov1998,BoykovICCV2001,BoykovEMMCVPR2001,Boykov2001,Kolmogorov2004} showed that min-cut solvers can  find the MAP solution of sub-modular energy functions of graphical models for binary segmentation. Approximation algorithms like swap-moves and $\alpha$-expansion~\cite{Boykov2001}  were developed to extend applicability of min-cut solvers to more than two labels. Semantic segmentation was further popularized by combining random forests with CRFs~\cite{ShottonECCV2006}. Recently, the performance on standard semantic segmentation benchmarks like Pascal VOC 2012~\cite{EveringhamIJCV2010} has been dramatically boosted by convolutional networks. Both deeper \cite{li2017not} and wider~\cite{noh2015learning, ronneberger2015u, yu2015multi} network architectures have been proposed. Advances like spatial pyramid pooling \cite{zhao2017pyramid} and atrous spatial pyramid pooling~\cite{chen2017rethinking} emerged to remedy limited receptive fields. Other approaches jointly train deep nets with CRFs~\cite{ChenSchwingICML2015,SchwingARXIV2015,GuistiICIP2013,SermanetICLR2014,LongCVPR2015,ChenICLR2015,ZhengICCV2015} to better capture the rich structure  present in natural scenes.

\noindent\textbf{CRF Inference:} 
Algorithmically, to find the MAP configuration, LP relaxations have been extensively studied in the 2000s~\cite{Schlesinger1976,Chekuri2001,Kolmogorov2005,Kolmogorov2006,Globerson2007,Werner2007,Johnson2008,Sontag2008,Jojic2010,Ravikumar2010,Werner2010,Meshi2011,Martins2011,Kappes2012,SchwingCVPR2011a,SchwingNIPS2012,SchwingICML2014,MeshiNIPS2015,MeshiNIPS2017}. 
Also, CRF inference was studied as a differentiable module within a deep net~\cite{zheng2015conditional,liu2015semantic, messaoud2018structural, graber2018deep, graber2019deep}. However, both directions remain computationally demanding, 
particularly if high order potentials are involved. 
We  therefore wonder whether recent progress in learning based combinatorial optimization yields effective algorithms for high order CRF inference in  semantic segmentation. 


\noindent\textbf{Learning-based Combinatorial Optimization:} Decades of research on combinatorial optimization, often also referred to as discrete optimization,  uncovered a large amount of valuable exact, approximation and heuristic algorithms. Already in the early 2000s, but more prominently recently~\cite{ZhangJAIR2000,BoyanJMLR2000,VinyalsNIPS2015,BelloARXIV2016,GuICLR2017,DaiNIPS2017}, learning based algorithms have been suggested  for combinatorial optimization. They are based on the intuition that instances of similar problems are often solved repeatedly. While humans have uncovered impressive heuristics, data driven techniques are likely to uncover even more compelling mechanisms. It is beyond the scope of this paper to review the vast literature on combinatorial optimization. Instead, we subsequently focus on learning based methods. Among the first is work by Boyan and Moore~\cite{BoyanJMLR2000}, discussing how to learn  to predict the outcome of a local search algorithm in order to bias future search trajectories. Around the same time, reinforcement learning techniques were used to solve resource-constrained scheduling tasks~\cite{ZhangJAIR2000}. Reinforcement learning is also the technique of choice for recent approaches addressing NP-hard tasks ~\cite{BelloARXIV2016,GuICLR2017,DaiNIPS2017,laterre2018ranked} like the traveling salesman, knapsack, maximum cut, and minimum vertex cover problems. Similarly, promising results exist for structured prediction problems like  dialog generation~\cite{li2016deep,williams2017hybrid,he2016dual}, program synthesis~\cite{bunel2018leveraging,liang2018memory,Pierrot2019LearningCN}, semantic parsing~\cite{liang2016neural}, architecture search~\cite{zoph2016neural}, chunking and parsing~\cite{sharaf2017structured}, machine translation~\cite{ranzato2015sequence,norouzi2016reward,bahdanau2016actor}, summarization~\cite{paulus2017deep}, image captioning~\cite{rennie2017self}, knowledge graph reasoning~\cite{xiong2017deeppath}, query rewriting~\cite{nogueira2017task,buck2017ask} and information extraction~\cite{narasimhan2016improving,qin2018robust}. Instead of directly learning to solve a given program, machine learning techniques have also been applied to parts of combinatorial solvers, \eg, to speed up branch-and-bound rules~\cite{Lagoudakis2001,SamulowitzAAAI2007,HeNIPS2014,KhalilAAAI2016}. 
We also want to highlight recent work on learning to optimize for continuous problems~\cite{LiICLR2017,AndrychowiczARXIV2016}. 

Given those impressive results on challenging real-world problems, we wonder: can we learn programs for solving higher order CRFs for semantic image segmentation? Since CRF inference is typically formulated as a combinatorial optimization problem, we want to know how  recent advances in learning based combinatorial optimization can be leveraged. 

%% file: app.tex
\begin{figure}
\centering
\includegraphics[width=8.5cm]{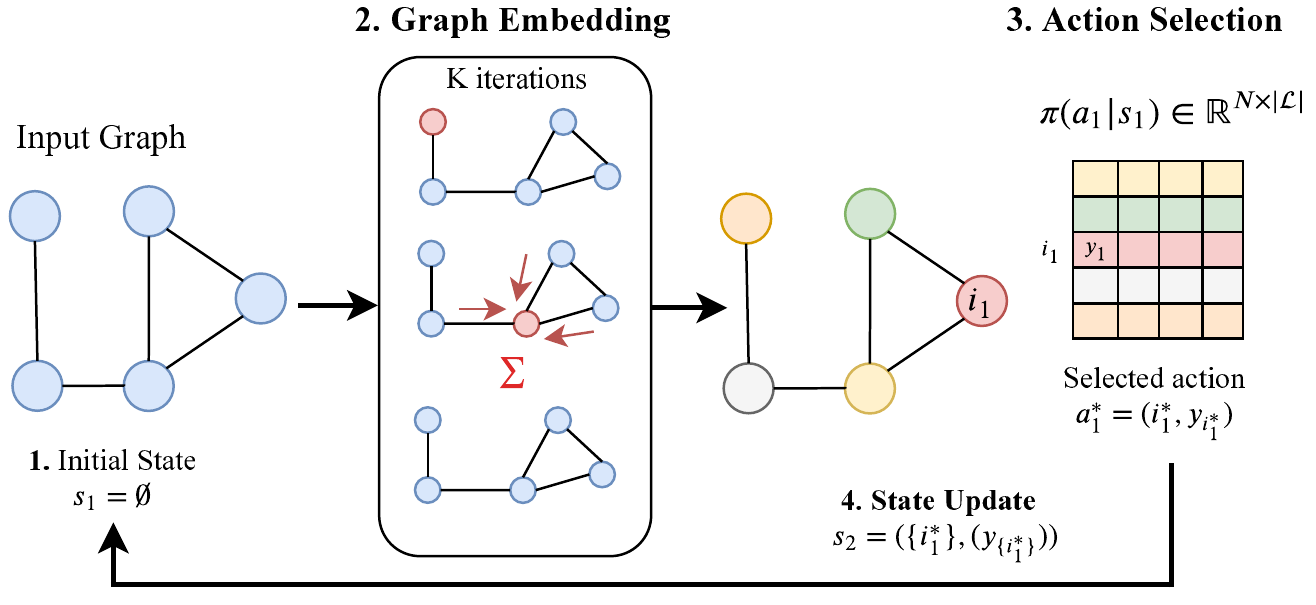}
\caption{ Illustration of one iteration of reinforcement learning for the inference task. The policy network samples an action $a_1=(i_1^\ast, y_{i_1^\ast})$ from the learned distribution $\pi(a_1|s_1) \in \mathbb{R}^{N\times|\mathcal{L}|}$ at iteration $t = 1$. } 
\label{fig:policynet}
\end{figure}

\section{Approach}
\label{sec:app}
We first present an overview of our approach before we discuss the individual components in greater detail.

\subsection{Overview}
Graphical models factorize a global energy function as a sum of local functions of two types: (1) local evidence; and (2) co-occurrence information. Both cues are typically obtained from deep net classifiers which are combined in a joint energy formulation. 
Finding the optimal semantic segmentation configuration, \ie, finding the minimizing argument of the energy, generally involves solving an NP-hard combinatorial optimization problem. Notable exceptions include  energies with sub-modular co-occurrence terms. 

Instead of using classical directions, \ie,  heuristics, exhaustive search, or relaxations, here, we assess  suitability of  learning based combinatorial optimization. Intuitively, we argue that CRF inference   for the task of semantic segmentation exhibits an inherent similarity which can be exploited by learning based algorithms. In spirit, this mimics the design of heuristic rules. However, different from hand-crafting those rules, we use a learning based approach. To the best of our knowledge, this is the first work to successfully apply learning based combinatorial optimization to CRF inference for semantic segmentation. We therefore first provide an overview of the developed approach, outlined in \figref{fig:Overview}. 

Just like classical approaches, we also use local evidence and co-occurrence information, obtained from deep nets. This information is consequently used to form an energy function defined over a Conditional Random Field (CRF). An example of a CRF with variables corresponding to superpixels (circles), pairwise potentials (edges) and higher order potentials obtained from object detections (fully connected cliques) is illustrated in \figref{fig:Overview}. However, different from classical methods, we find the minimizing configuration of the energy by repeatedly applying a learned policy network. In every iteration, the policy network selects a random variable, \ie, the pixel and its label by computing a probability distribution over all currently unlabeled pixels and their labels. Specifically, the pixel and label are determined by choosing the highest scoring entry in a matrix where the number of rows and columns correspond to the currently unlabeled pixels and the available labels respectively, as illustrated in \figref{fig:policynet}.


\subsection{Problem Formulation}
Formally, given an image $x$, we are interested in predicting the semantic segmentation $y = (y_1, \ldots, y_N) \in \cY$. Hereby, $N$ denotes the total number of pixels or superpixels, and the semantic segmentation of a superpixel $i\in\{1, \ldots, N\}$ is referred to via $y_i\in\cL = \{1, \ldots, |\cL|\}$, which can be assigned one out of $|\cL|$ possible discrete labels from the set of possible labels $\cL$. The output space is denoted $\cY = \cL^N$.

Classical techniques obtain local evidence $f_i(y_i)$ for every pixel or superpixel, and co-occurrence information in the form of pairwise potentials $f_{ij}(y_i, y_j)$ and higher order potentials $f_{c}({y}_{c})$. The latter assigns an energy to a clique $c\subseteq\{1, \ldots, N\}$ of variables $y_c = (y_i)_{i\in c}$. For readability, we  drop the dependence of the energies  $f_i$, $f_{ij}$ and $f_{c}$ on the image $x$ and the parameters of the employed deep nets. The goal of energy based semantic segmentation is to find the configuration $y^\ast$ which has the lowest energy $E(y)$, \ie, 
\be
\begin{aligned}
y^\ast \!=\!  \arg\min_{y\in\cY} E(y) \!\triangleq\! \sum_{i=1}^N f_i(y_i) +\!\!\!\!\sum_{(i,j)\in\cE}\!\!\!\! f_{ij}(y_i, y_j) + \!\sum_{c\in\cC}\! f_{c}( y_{c}).
\end{aligned}
\label{eq:energy}
\ee
Hereby, the sets $\cE$ and  $\cC$ subsume respectively the captured set of pairwise and higher order co-occurrence patterns. Details about the potentials are presented in  \secref{sec:energy}.

Solving the combinatorial program given in \equref{eq:energy}, \ie, inferring the optimal configuration $y^\ast$ is generally NP-hard. 
Different from existing methods, we develop a learning based combinatorial optimization heuristic for semantic segmentation with the intention to better capture the intricacies of energy minimization than can be done by hand-crafting rules. The developed heuristic sequentially labels one variable $y_i$, $i\in \{1, \ldots, N\}$, at a time.

Formally, selection of one superpixel at a time can be formulated in a reinforcement learning context, as shown in \figref{fig:policynet}. Specifically, an agent operates in $t\in\{1, \ldots, N\}$ time-steps according to a policy $\pi(a_t | s_t)$ which encodes a probability distribution over actions $a_t\in\cA_t$ given the current state $s_t$. 
The current state subsumes in selection order the indices of all currently labeled variables $I_t\subseteq\{1, \ldots, N\}$  as well as their labels $y_{I_t} = (y_i)_{i\in I_t}$, \ie, $s_t \in \{ (I_t, y_{I_t}) : I_t \subseteq \{1, \ldots, N\}, y_{I_t} \in \cL^{|I_t|}\}$.
We start with $s_1 = \emptyset$. The set of possible actions $\cA_t$ is  the concatenation of the label spaces $\cL$ for all currently unlabeled pixels $j\in\{1, \ldots, N\}\setminus I_t$, \ie, $\cA_t = \bigoplus_{j\in\{1, \ldots, N\}\setminus I_t} \cL$. We emphasize the  difference between the concatenation operator and the product operator used to obtain the semantic segmentation output space $\cY = \cL^N$, \ie, the proposed approach does not operate in the product space. 

As mentioned before, the policy $\pi(a_t|s_t)$ results in a probability distribution over actions $a_t\in\cA_t$ from which we greedily select the most probable action 
$$
a_t^\ast = \arg\max_{a_t \in \cA_t} \pi(a_t | s_t).
$$
The most probable action $a_t^\ast$ can be decomposed into the index for the selected variable, \ie, $i^\ast_t$ and its state $y_{i^\ast_t}\in\cL$. We obtain the subsequent state $s_{t+1}$ by combining the extracted variable index $i^\ast_t$ and its labeling with the previous state $s_t$. Specifically, we obtain $s_{t+1} = s_t \oplus (i^\ast_t, y_{i^\ast_t})$ by slightly abusing the $\oplus$-operator to mean concatenation to a set and a  list maintained within a state. 

\begin{figure}[t]
\makeatletter
\newcommand{\removelatexerror}{\let\@latex@error\@gobble}
\makeatother
\centering
\begin{minipage}{\linewidth}
\begingroup
\removelatexerror
\begin{algorithm}[H]
   \caption{Inference Procedure}
\centering
\begin{algorithmic}[1]
   \STATE $s_1 = \emptyset$; 
   \FOR {$t=1$ {\bfseries to} $N$}
   		\STATE $a_t^\ast = \arg\max_{a_t\in\cA_t} \pi(a_t|s_t)$
		\STATE $(i^\ast_t, y_{i^\ast_t}) \leftarrow a_t^\ast$
		\STATE $s_{t+1} = s_t \oplus (i_t^\ast, y_{i^\ast_t})$
   \ENDFOR
   \STATE {\bfseries Return:} $\hat y \leftarrow s_{N+1}$
\end{algorithmic}
   \label{alg:inf}
\end{algorithm}
\endgroup
\end{minipage}
\vspace{-0.75cm}
\end{figure}

Formally, we summarize the developed reinforcement learning based semantic segmentation algorithm used for inferring a labeling $\hat y$ in \algref{alg:inf}. 
In the following, we  describe the policy function $\pi_\theta(a_t|s_t)$, which we found to work well for semantic segmentation, and different variants to learn its parameters $\theta$. 


\input{tab_reward_energy}

\subsection{Policy Function}
\label{sec:policynet}

We model the policy function $\pi_\theta(a_t|s_t)$  using a graph embedding network~\cite{dai2016discriminative}. The input to the network is a weighted graph $G(V,\cE, w)$, where nodes $V = \{1, \ldots, N\}$, correspond to variables, \ie, in our case superpixels,  $\cE$ is a set of edges connecting neighboring superpixels, as illustrated in \figref{fig:Overview} and $w:\cE\rightarrow\mathbb{R}^+$ is the edge weight function. 
The weights $\{w(i,j)\}_{\{j:(i,j)\in \cE\}}$ on the edges between a given node $i$ and its neighbors $\{j:(i,j)\in \cE\}$ form a distribution, obtained by normalizing the dot product between the hypercolumns~\cite{hariharan2015hypercolumns} $g_i$ and $g_j$ via a softmax across neighbors. 
At every iteration, the state $s_t$ is encoded in the graph $G$ by tagging node $i\in V$ with a scalar $h_i=1$ if the node is part of the already labeled set $I_t$, \ie, if $i\in I_t$ and $0$ otherwise. Moreover, a one-hot encoding $\tilde{y}_{i} \in \{0,1\}^{|\cL|}$  encodes the selected label of nodes $i\in I_t$.   We set $\tilde{y}_{i}$ to equal the all zeros vector if node $i$ has not been selected yet. 

Every node $i\in V$ is represented by a $p$-dimensional embedding, where $p$ is a hyperparameter. The embedding is composed of $\tilde{y}_{i}$, $h_{i}$ as well as superpixel features $ b_{i} \in \mathbb{R}^F$ which encode appearance and bounding box characteristics that we discuss in detail in \secref{sec:exp}. 

The output of the network is a $|\cL|$-dimensional vector $\pi_i$ for each node $i\in V$, representing the scores of the $|\cL|$ different labels for variable $i$.  

The network iteratively generates a new representation $\mu_{i}^{(k+1)}$ for every node $i\in V$ by aggregating the current embeddings $\mu_{i}^{(k)}$ according to the graph structure $\cE$ starting from $\mu_{i}^{(0)} = {\mathbf 0}$, $\forall i\in V$. After $K$ steps, the embedding  captures long range interactions between the graph features as well as the graph properties necessary to minimize the energy function $E$. Formally, the update rule for  node $i$ is  
\be
\mu_{i}^{(k+1)}\! \leftarrow\! \text{Relu}\!\left(\theta_{1}^{(k)}  h_{i}\!+\!\theta_{2}^{(k)}\!  \tilde{y}_{i} \! +\! \theta_{3}^{(k)} b_{i}\!+\!\theta_{4}^{(k)}\!\!\!\!\! \sum_{j : (i,j)\in\cE}\!\!\!\!\!{w(i,j) \mu_{j}^{(k)}  }\! \right) , 
  \label{eq:gnn}
\ee
where $ \theta_{1}^{(k)}\! \in\! \mathbb{R}^{p}$, $ \theta_{2}^{(k)}\! \in\! \mathbb{R}^{p \times |\mathcal{L}|}$ , $ \theta_{3}^{(k)}\!  \in\! \mathbb{R}^{p \times F}$ and $ \theta_{4}^{(k)}\! \in\! \mathbb{R}^{p\times p}$ are trainable parameters. After  $K$ steps, $\pi_i$ for every unlabeled node $i \in \{1, \ldots, N\}\setminus I_t$ is obtained via
\be
\pi_i = \theta_{5} \mu_{i}^{(K)} \quad\forall i \in \{1, \ldots, N\}\setminus I_t,
\ee
where $\theta_{5} \in \mathbb{R}^{|\mathcal{L}|\times p}$ is another trainable model parameter. 
We illustrate the policy function $\pi_\theta(a_t|s_t)$ and one iteration of inference in \figref{fig:policynet}. 

\subsection{Reward Function:} 
To train the policy, ideally, the reward function $r_{t}(s_{t}, a_{t})$ is designed such that the cumulative reward coincides exactly with the objective function that we aim at maximizing, \ie, $\sum_{t=1}^{N} r_{t}(s_{t}, a_{t})=-E(\hat y)$, where $\hat y$ is extracted from $s_{N+1}$. Hence, at step $t$, we define the reward as the difference between the value of the negative new energy $E_{t}$ and the negative energy from the previous step $E_{t-1}$, \ie, $r_{t}(s_{t},a_{t})=E_{t-1}(y_{I_{t-1}})-E_{t}(y_{I_{t}})$, where $E_{0}=0$. Potentials depending on variables that are not labeled  at time $t$ are not incorporated in the evaluation of $E_{t}(y_{I_t})$.

We also study a second scheme, where the reward is truncated to $+1$ or $-1$, \ie, $r_{t}(s_t,a_t) \in \{-1,1\}$. For every selected node $i_{t}$, with label $y_{i_{t}}$, we compare the energy function $E_{t}(y_{I_t})$ with the one obtained when using all  other labels $\hat y_{i_t} \in \cL\setminus y_{i_t}$. 
If the chosen label $y_{i_{t}}$ results in the lowest energy, the obtained reward is $+1$, otherwise it is $-1$.

Note that the unary 
potentials 
result in a reward for every time step. Pairwise and high order potentials result in a sparse reward as their value is only available once all the superpixels forming the pair or clique are labeled. We illustrate the energy and reward computation on a graph with three fully connected nodes in \tabref{tab:reward_energy}. 

\subsection{Learning Policy Parameters}
\label{sec:policy_parameters}
To learn the parameters $\theta$ of the policy function $\pi_\theta(a_t|s_t)$, a plethora of reinforcement learning algorithms are applicable. To provide a careful assessment of the developed approach, we study  two different techniques, Q-learning and Monte-Carlo Tree Search, both of which we describe next. 


\noindent\textbf{Q-learning:} 
In the context of Q-learning, we interpret the $|\cL|$-dimensional policy network output vector  corresponding to a currently unlabeled node $i \in \{1, \ldots, N\}\setminus I_t$ as the Q-values $Q(s_{t}, a_{t};  \theta)$ associated to the action $a_{t}$ of selecting  node $i$ and assigning label $y_i \in \cL$. Since we only consider actions to label currently unlabeled nodes we obtain a total of $|\cA_t|$ different Q-values.

We perform standard Q-learning and minimize the squared loss $(z-Q(s_{t}, a_{t};  \theta))^{2}$, where we use target $z = \gamma \max_{a'}Q(s_{t+1}, a'; \theta)+r_{t}(s_{t},a_{t})$ for a non-terminal state. The reward is denoted $r_{t}$ and detailed above. 
The terminal state is reached when all the nodes are labeled. 

Instead of updating the Q-function based on the current sample ($s_{t}, a_{t}, r_{t}(s_{t},a_{t}),s_{t+1}$), we use a replay memory populated with samples (graphs) from previous episodes. In every iteration, we select a batch of samples and perform stochastic gradient descent on the squared loss. 

During the exploration phase, beyond random actions, we encourage the following three different sets of actions to generate more informative rewards for training: (1) $\mathcal{M}_{1}$: Choosing nodes that are adjacent to the already selected ones in the graph. Otherwise, the reward will only be based on the unary terms as the pairwise term is only evaluated if the neighbors are labeled ($t=2$ in \tabref{tab:reward_energy}); (2) $\mathcal{M}_{2}$: Selecting nodes with the lowest unary distribution entropy. The low entropy indicates a high confidence of the unary deep net. Hence, the labels of the corresponding nodes are more likely to be correct and provide useful information to neighbors with higher entropy in the upcoming iterations. (3) $\mathcal{M}_{3}$: Assigning the same label to nodes forming the same higher order potential. Further details are in \textbf{Appendix B}.
\input{results_tab}

\noindent\textbf{Monte-Carlo Tree Search:} 
While DQN tries to learn a policy from looking at samples representing one action at a time, MCTS has the inherent ability to update its policy after looking multiple steps ahead via a tree search procedure. 
At training time, through extensive simulations, MCTS builds a \textit{statistics tree}, reflecting an empirical distribution $\pi^\text{MCTS}(a_{t}|s_{t})$. Specifically, for a given image, a node in the search tree corresponds to the state $s_{t}$ in our formulation and an edge corresponds to a possible action $a_{t}$. The root node is initialized to $s_{1}=\emptyset$. The statistics stored at every node correspond to (1) $N(s_t)$: the number of times state $s_t$ has been reached, (2) $N(a_t|s_t)$: the number of times action $a_t$ was chosen in state $s_t$ in all previous simulations, as well as (3) $\tilde{r}_{t}(s_t,a_t)$: the averaged reward across all simulations starting at $s_{t}$ and taking action $a_{t}$. The MCTS policy is defined as  $\pi^\text{MCTS}(a_t|s_t)=\frac{N(a_t|s_t)}{N(s_t)}$. The simulations follow an exploration-exploitation procedure modeled by a variant of the Probabilistic Upper Confidence Bound (PUCB) \cite{silver2016mastering}: $U(a_{t},s_{t})  = \frac{\tilde{r}_{t}(s_t,a_t)}{N(a_t|s_t)} +  \pi_\theta(a_t|s_t) \frac{\sqrt{N(s_t)}}{1 + N(a_t|s_t)}$. During exploration, we additionally encourage the same action sets $\mathcal{M}_{1}$, $\mathcal{M}_{2}$ and $\mathcal{M}_{3}$ used for DQN.
Also, similarly to DQN, the generated experiences $(s_{t}, \pi^\text{MCTS})$ are stored in a replay buffer. The policy network is then trained through a cross entropy loss 
    \be
L(\theta) = - \sum_{s}  \sum_{a}  \pi^\text{MCTS}(a|s) \log\pi_\theta(a|s).
\label{eq:lossf}
\ee
to mach the empirically constructed distribution. Here, the second sum is over all valid actions from a state $s$ sampled from the replay buffer and $\pi^\text{MCTS}(a|s)$ is the corresponding empirically estimated distribution.

A more detailed description of the MCTS search process, including pseudo-code is available in \textbf{Appendix B}. At inference time, we use a low budget of simulations. Final actions are taken according to the constructed $\pi^\text{MCTS}(a_{t}|s_{t})$. 

The replay-memory for both DQN and MCTS is divided into two chunks. The first chunk corresponds to the unary potential, while the second chunk corresponds to the overall energy function. A node is assigned to the second chunk if its associated reward is higher than the one obtained from its unary labeling.  This ensures positive rewards from all the potentials during training. Every chunk is further divided into $|\cL|$ categories corresponding to the $|\cL|$  classes of the selected node. This guarantees a balanced sampling of the label classes in every batch during training. Beyond DQN and MCTS, we experimented with policy gradients but could not get it to work as it is an on-policy algorithm.  Reusing experiences for the structured replay buffer was crucial for success of the learning algorithm.

\begin{figure*}[!t]
\centering
\includegraphics[width=15cm]{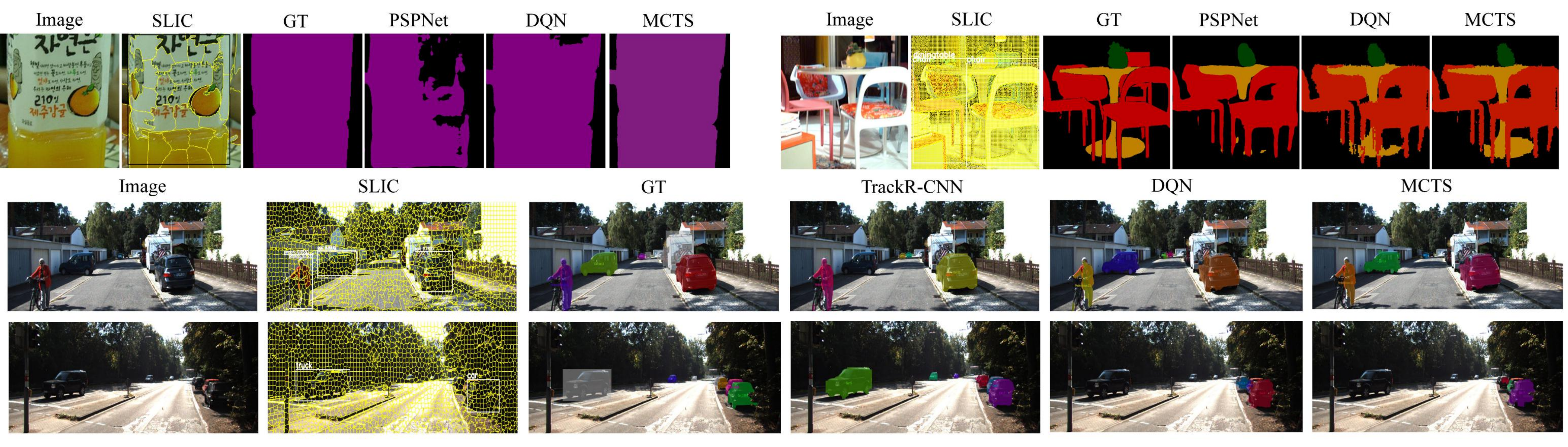}
\vspace{-0.3cm}
\caption{Success cases.}
\label{fig:success_cases}
\end{figure*}

\subsection{Energy Function}
\label{sec:energy}

Finally we provide details on the energy function $E$ given in \equref{eq:energy}. The unary potentials $f_i(y_i) \in \Bbb{R}^{ |\cL|  } $ are obtained from a semantic segmentation deep net. The pairwise potential encodes smoothness and is computed as follows:
\be
f_{i,j}(y_{i},y_{j}) = \psi(y_{i},y_{j})  \cdot \alpha_{p} \cdot   \mathbbm{1}_{ |g_{i}^{T}g_{j}|<\beta_{p}  },
\ee
where $\psi(y_{i},y_{j})$ is the label compatibility function describing the  co-occurrence of two  classes in adjacent locations and is given by the Potts model:
\begin{equation}
   \psi(y_{i},y_{j})= \begin{cases}
  1 &  \text{if $y_{i} \neq y_{j}$ }\\
  0& \text{otherwise}
  \end{cases}.
\end{equation}
Moreover, $|g_{i}^{T}g_{j}|$ is the above defined unnormalized weight $w(i,j)$ for the edge connecting the $i^\text{th}$ and $j^\text{th}$ nodes, \ie, superpixels. Intuitively, if the dot product between the hypercolumns $g_{i}$ and $g_{j}$ is smaller than a threshold $\beta_{p}$ and the two superpixels are labeled differently, a penalty of value $\alpha_{p}$ incurs. 

While the pairwise term mitigates boundary errors, we address recognition errors with two detection-based higher order potentials~\cite{arnab2016higher}. For this purpose, we use  the YoloV2 bounding box object detector~\cite{redmon2016you} as it ensures a good tradeoff between speed and accuracy. Every bounding box $b$ is presented by a tuple $(l_{b},c_{b},I_{b})$, where $l_{b}$ is the class label of the detected object, $c_{b}$ is the confidence score of the detection and $I_{b} \subseteq \{1, \ldots, N\}$ is the set of superpixels that belong to the foreground detection  obtained via Grab-Cut~\cite{rother2004grabcut}. 

The first higher order potential (HOP1) encourages superpixels within a bounding box to take the bounding box label, while enabling  recovery from false detections that do not agree with other energy types.  For this purpose, we add an auxiliary variable $z_{b}$ for every bounding box $b$.
We use $z_b=1$, if the bounding box is inferred to be valid, otherwise $z_b = 0$.
Formally,  
\begin{equation}
  f(y_{I_b},z_{b})=\begin{cases}
   w_{b} \cdot c_{b} \cdot  \sum_{i\in I_b}  \mathbbm{1}_{y_{i} = l_{b} }    & \text{if $ z_{b} =0$ }\\
   w_{b} \cdot c_{b} \cdot \sum_{i\in I_b}  \mathbbm{1}_{y_{i} \neq l_{b}}   & \text{if $z_{b} =1$}
  \end{cases},
\end{equation}
where, $w_{b}\in\mathbb{R}$ is a weight parameter. This potential can be simplified into a sum of pairwise potentials between $z_{b}$ and each $y_{i}$ with $i\in I_b$, \ie,  $f(y_{I_b},z_{b})  = \sum_{i\in I_b} f_{i,b}(  y_{i} , z_{b}  ) $, where:
\begin{equation}
  f_{i,b}(y_{i},z_{b})=\begin{cases}
   w_{b}  \cdot c_{b}  \cdot \mathbbm{1}_{y_{i}=l_{b} }    & \text{if $z_{b} = 0$ }\\
   w_{b}  \cdot c_{b}  \cdot  \mathbbm{1}_{y_{i} \neq l_{b} }   & \text{if $z_{b} = 1$ }
  \end{cases}.
\end{equation}
This simplification enables solving the higher order potential using traditional techniques like mean field inference \cite{arnab2016higher}.

To show the merit of the RL framework, we introduce another higher order potential (HOP2) that can not be seamlessly reduced to a pairwise one:
\begin{equation}
  f(y_{I_b} )= \lambda_{b} \cdot \mathbbm{1}_{ (\sum_{i\in I_b}{ y_{i} = l  })< \frac{|I_{b}|}{C}  } ,
\end{equation} 
with $\lambda_{b}$ and $C$ being scalar parameters. This potential is evaluated for bounding boxes with special characteristics to encourage the superpixels in the bounding box to be of label $l$. Intuitively, if the number of superpixels  $i\in I_{b}$ having  label $l$ is less than a threshold $\frac{|I_{b}|}{C}$, a penalty $\lambda_{b}$ incurs. For Pascal VOC, we evaluate the potential on bounding boxes $b$  included in larger bounding boxes, as we noticed that the unaries frequently miss small objects overlapping with other larger objects in the image ($l=l_{b}$). For MOTS, we evaluate this potential on bounding boxes of type `pedestrians' overlapping with bounding boxes of type `bicycle.' As cyclists should not be labeled as pedestrians, we set $l$ to be the background class.
Transforming this term into a pairwise one to enable using traditional inference techniques requires an exponential number of auxiliary variables. 

%% file: tab_reward_energy.tex
\begin{table*}[!t]
\centering
{\scriptsize
\centering
\caption{\footnotesize{ Illustration of the energy reward computation following the two proposed reward schemes on a fully connected graph with 3 nodes.}}
\vspace{-0.3cm}
\label{tab:reward_energy}
\setlength\tabcolsep{3pt}
\begin{tabular}{cccccccccc} \toprule
   {\scriptsize  $t$}   &  {\scriptsize $i_{t}$} &  {\scriptsize $E_{t}$ }  & {\scriptsize  $r_{t}=-(E_{t}-E_{t-1})$ }  &   {\scriptsize$r_{t}=\pm 1$} & {\scriptsize Graph} \\      \midrule           
      \textbf{0}      &    $-$   &     0                     &       $-$      &    $-$   &  \includegraphics[width=1.2cm]{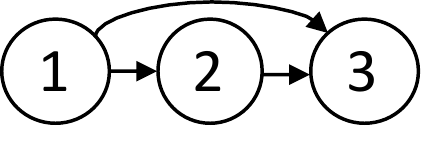} \\        
      \textbf{1}     &     1    &   $f_{1}(y_{1})$  & $-f_{1}(y_{1})$  &  $-\!1 \!+\! 2\! \cdot \mathbb{1}_{ \{ (E_{t}(y_{1}) < E_{t}(\hat y_{1} )) \forall \hat y_{1}  \} } $ &  \includegraphics[width=1.2cm]{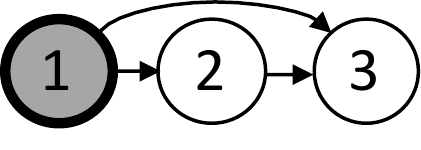}   \\
     \textbf{2}      &     2    &   $f_{1}(y_{1})\!+\!f_{2}(y_{2})\!+\! f_{12}(y_{1},y_{2})$  & $-f_{2}(y_{2})\!-\!f_{12}(y_{1},y_{2})$ & $-\!1\! +\! 2\! \cdot \mathbb{1}_{ \{(E_{t}(y_{1},y_{2}) < E_{t}(y_{1}, \hat y_{2}))  \forall \hat y_{2}  \}} $ &  \includegraphics[width=1.2cm]{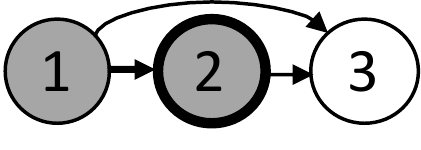}  \\
     \textbf{3}      &    3     &  $f_{1}(y_{1})\!+\!f_{2}(y_{2})+f_{3}(y_{3})+\! f_{12}(y_{1},y_{2})\!$      &  $-f_{3}(y_{3})\!-\! f_{23}(y_{2},y_{3})$ &  $-\!1\! +\! 2\! \cdot \mathbb{1}_{ \{(E_{t}(y_{1},y_{2}, y_{3}) < E_{t}(y_{1}, y_{2}, \hat  y_{3}))  \forall \hat y_{3} \}} $  &  \includegraphics[width=1.2cm]{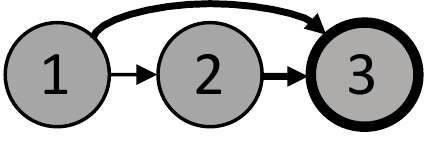} \\
             &          & $+\!f_{23}(y_{2},y_{3})+\! f_{13}(y_{1},y_{3})\!+\!f_{ \{1,2,3\} }(y_{1},y_{2},y_{3})$     &$ -\! f_{13}(y_{1},y_{3})-\!f_{ \{1,2,3\} }(y_{1},y_{2},y_{3})$     &  \\

             \midrule
\bottomrule
\end{tabular}}
\vspace{-0.3cm}

\end{table*}

%% file: results_tab.tex
\begin{table*}[!t]
\centering
{\tiny
\centering
\caption{\footnotesize{Performance results for the minimizing the energy function $E_{t}$ under reward scheme 1 ($R^{1}_{t}=-(E_{t}-E_{t-1})$) and reward scheme 2 ($R^{2}_{t}=\pm1$). }}
\vspace{-0.3cm}
\label{tab:Unary_Pairwise_Potential}
\setlength\tabcolsep{1.7pt}
\begin{tabular}{cc|c|c|c|ccccc|ccccccccc|ccccccccc|cccc} \toprule
&&\multirow{0}{*}{\rotatebox[origin=c]{90}{ \textbf{Nodes}}}  & \multirow{0}{*}{\rotatebox[origin=c]{90}{ \textbf{Metrics}}}  &\multirow{-0.4}{*}{\rotatebox[origin=c]{90}{ \textbf{Supervised}}}  &\multicolumn{5}{c|}{\textbf{Unary} }  & \multicolumn{9}{c|}{ \textbf{Unary + Pairwise}} & \multicolumn{9}{c|}{\textbf{Unary + Pairwise + HOP1} }  & \multicolumn{4}{c}{ \textbf{Unary + Pairwise + HOP1 + HOP2}} \\ 
\cmidrule(lr){6-10} \cmidrule(lr){11-19}  \cmidrule(lr){20-28} \cmidrule(lr){29-32}

&&  & &  &  \textbf{BP}  & \multicolumn{2}{c}{ \textbf{$R^{1}_{t}$} }&  \multicolumn{2}{c|}{\textbf{ $R^{2}_{t}$} } &\textbf{BP}  & \textbf{TBP}  & \textbf{DD} & \textbf{L-Flip} & \textbf{$\alpha$-Exp} & \multicolumn{2}{c}{\textbf{$R^{1}_{t}$}} &  \multicolumn{2}{c|}{\textbf{$R^{2}_{t}$}}  &   \textbf{BP}  &\textbf{TBP}  & \textbf{DD} &\textbf{ L-Flip} & \textbf{$\alpha$-Exp}   & \multicolumn{2}{c}{\textbf{$R^{1}_{t} $ }}&  \multicolumn{2}{c|}{ \textbf{$R^{2}_{t}$}}   & \multicolumn{2}{c}{\textbf{$R^{1}_{t}$} }&  \multicolumn{2}{c}{\textbf{ $R^{2}_{t}$} }\\ \cmidrule(lr){7-8}  \cmidrule(lr){9-10}   \cmidrule(lr){16-17}  \cmidrule(lr){18-19} \cmidrule(lr){25-26} \cmidrule(lr){27-28} \cmidrule(lr){29-30} \cmidrule(lr){31-32} 

&&&  & &  & \textbf{DQN}  &  \textbf{MCTS}  &  \textbf{DQN}   &   \textbf{MCTS}   &    &  &  &  & &   \textbf{DQN}  &  \textbf{MCTS}   & \textbf{DQN}  & \textbf{MCTS}  &   &  &  &  &     &  \textbf{DQN}  &  \textbf{MCTS}  &  \textbf{DQN}  &   \textbf{MCTS}  &  \textbf{DQN} &   \textbf{MCTS}  & \textbf{DQN } & \textbf{MCTS} \\ \midrule  

&&\multirow{-0.5}{*}{\rotatebox[origin=c]{90}{\textbf{50}}} & \textbf{ IoU \tiny{(sp)}} & 85.21 &  \textbf{88.59}	& 88.04 &88.19 &	\textbf{88.59}  &	\textbf{88.59}	&	\textbf{88.73} &	\textbf{88.73}	 & \textbf{88.73} &	\textbf{88.73} & 88.72 &	43.31 &	66.51 &	87.91  &	\textbf{88.73}	&	89.26 &	89.27	& 89.27&	89.27	& 88.58	& 57.43 &		73.37&	89.55 & \textbf{89.66}&	58.34	&	73.85& 90.05& \textbf{90.09} \\

\multirow{-0}{*}{\rotatebox[origin=c]{90}{ \textbf{Pascal VOC}}}&&& \textbf{IoU \tiny{(p)}} & 69.05  &  \textbf{72.56}	& 70.77& 71.99 &	\textbf{72.56}	&	\textbf{72.56}	&	\textbf{72.43} &	\textbf{72.43} &	\textbf{72.43} &	\textbf{72.43} 	& \textbf{72.43} &	38.54&38.75	& 72.16	&\textbf{72.43}	&	72.59 &	72.59 &	72.59 &	72.60&	72.35	& 51.88	& 53.71 &72.83	&	\textbf{72.85}&	50.53&51.71  &	72.94 &  \textbf{72.95} \\  \cmidrule(lr){2-32}

&&\multirow{-0.9}{*}{\rotatebox[origin=c]{90}{\textbf{250}}}  &  \textbf{IoU \tiny{(sp)}} & 83.54 &   \textbf{88.01}	& 87.29	& \textbf{88.01}& \textbf{88.01}	& 	\textbf{88.01}	& 	88.10	& 88.10 & 	88.10	& 88.10 & 	88.10	& 88.06& 88.22	& \textbf{88.56}	& 	 88.52	&	88.54	& 88.53 & 	88.55	& 88.54	& 88.07	& 60.60& 64.82	& \textbf{88.94}	 & 88.91 	& 	82.19& 81.89 	& 89.30  & \textbf{89.57}	   \\

&&&\textbf{ IoU \tiny{(p)}} & 75.88  & 80.64&	80.47&	\textbf{80.64}	& \textbf{80.64}	& \textbf{80.64}&	80.68	&80.68 &	80.68	& 80.68	& 80.68&	80.54 &80.86&	\textbf{80.84} 	& 80.75 &	80.91	& 80.91 &	80.93 &	80.91	& 80.65	& 	57.36 & 59.73 &	\textbf{81.07} &  81.05	&	 74.94	&	74.77   & 81.23	& \textbf{81.33}  \\ \cmidrule(lr){2-32}

&&\multirow{-0.9}{*}{\rotatebox[origin=c]{90}{\textbf{500}}}&  \textbf{IoU \tiny{(sp)}}& 84.91  & \textbf{87.39}	& 87.34& \textbf{87.39}	& \textbf{87.39}	 &	\textbf{87.39}		& 87.55	& 87.55	& 87.56 &	87.55	&87.55& 82.23	& 83.67  & 87.80&	 \textbf{87.84}	&		87.95	& 87.96	& 87.96	& 87.95	& 87.54&	37.80	& 57.66  & \textbf{88.73}		& 88.69  & 43.99&	45.67	&\textbf{88.43}&	88.21 \\

&&& \textbf{IoU \tiny{(p)}}  & 77.93 &\textbf{82.35}	&82.20 &	\textbf{82.35} &	\textbf{82.35} &	\textbf{82.35}	&	82.48 &	82.48 &	82.48 &	82.47 &	82.47 &	77.36 & 79.14 &	82.64  &	\textbf{82.70}	 & 82.72 &	82.72 & 82.72&	82.71 & 82.47 & 36.65 & 52.73 & \textbf{83.05} &	 82.95		& 41.74 & 42.44  & \textbf{82.79} &	 82.67  \\ 
  
\bottomrule

\multirow{-1}{*}{\rotatebox[origin=c]{90}{ \textbf{MOTS}}}&&\multirow{-0.9}{*}{\rotatebox[origin=c]{90}{\textbf{2000}}}& \textbf{ IoU \tiny{(sp)}} & 82.49 & 82.64 & 80.98  &  	82.64&82.64 & 82.64&	 82.64 &	82.64	&	82.64 &	82.64	 &  	82.64	 & 80.39   & 82.64 &	82.65  &	\textbf{82.64} &	83.17  &	83.17&	 83.17&	83.17	& 83.16 &83.14		&83.30 	& 	83.27	& \textbf{83.28}	& 83.13  &	83.19	& \textbf{83.29}	& \textbf{ 83.29} \\

&&&\textbf{ IoU \tiny{(p)}} &  79.01& 79.23 & 73.85  &	 79.82&	79.85	& 79.85 & 	 79.86  &	79.86	&	79.86 &		79.86 &  	79.86	 & 78.08 & 78.85 &	 79.88 &\textbf{ 79.86} 	&  81.21 	& 81.21	 &	81.21 & 81.21 & 81.17 & 80.61	& 81.92 & 82.68   & \textbf{82.69}	& 80.61&	80.63& \textbf{82.77}  & \textbf{82.77}	 \\  
\bottomrule
\end{tabular}}
\vspace{-0.25cm}
\end{table*}

%% file: exp.tex
\section{Experiments}
\label{sec:exp}
In the following, we evaluate our learning based inference algorithm on Pascal VOC~\cite{EveringhamIJCV2010} and MOTS~\cite{EveringhamIJCV2010} datasets. The original Pascal VOC dataset contains 1464 training and 1449 validation images. In addition to this data, we make use of the annotations provided by~\cite{BharathICCV2011}, resulting in a total of 10582 training instances. The number of classes is 21. MOTS is a multi-object tracking and segmentation dataset for cars and pedestrians (2 classes). It consists of 12 training sequences (5027 frames) and 9 validation ones (2981 frames). In this work, we perform semantic segmentation at the level of superpixels, generated using SLIC~\cite{achanta2010slic}. Every superpixel corresponds to a node $i$ in the graph as illustrated in \figref{fig:Overview}. The unary potentials at the pixel level are obtained from PSPNet~\cite{zhao2017pyramid} for Pascal VOC and TrackR-CNN~\cite{voigtlaender2019mots} for MOTS. The superpixels' unaries are the average of the unaries of all the pixels that belong to that superpixel. The higher order potential is based on the YoloV2~\cite{redmon2016you} bounding box detector. Additional training and implementation details are described in \textbf{Appendix A}.

\noindent\textbf{Evaluation Metrics:} 
As evaluation metrics, we use intersection over union (IoU) computed at the level of both superpixels (sp) and pixels (p). IoU (p) is obtained after mapping the superpixel level labels to the corresponding set of pixels. 

\noindent\textbf{Baselines:} 
We compare our results to the segmentations obtained by five different solvers from three categories: (1) message passing algorithms, \ie, Belief propagation (BP)~\cite{pearl1982reverend} and Tree-reweighted Belief Propagation (TBP)~\cite{wainwrighttree}, (2) a Lagrangian relaxation method, \ie, Dual Decomposition Subgradient (DD)~\cite{kappes2012bundle}, and (3) move making algorithms, \ie,  Lazy Flipper (LFlip)~\cite{andres2012lazy} and $\alpha$-expansion as implemented in~\cite{fix2011graph}. Note that these solvers can not optimize our HOP2 potential. Besides, we train a supervised model that predicts the node label from the provided node features.

\noindent\textbf{Performance Evaluation:} 
We show the results of solving the program given in \equref{eq:energy} in  \tabref{tab:Unary_Pairwise_Potential}, for  unary (Col.~5), unary plus pairwise (Col.~6), unary plus pairwise plus HOP1 (Col.~7) and unary plus pairwise plus HOP1 and HOP2 (Col.~8) potentials. For every potential type, we report results on graphs with superpixel numbers  50, 250 and 500 for Pascal VOC and 2000 for MOTS, obtained from DQN and MCTS, trained each with the two reward schemes discussed in \secref{sec:policy_parameters}. Since MOTS has small objects, we opt for a higher number of superpixels. 
It is remarkable to observe that DQN and MCTS are able to learn heuristics which outperform the baselines. 
Interestingly, the policy has learned to produce better semantic segmentations than the ones obtained via  MRF energy minimization.  Guided by a reward derived just from the energy function, the graph neural network (the policy) learns characteristic node embeddings for every semantic class by leveraging the hypercolumn and bounding box features as well as the neighborhood characteristics. 
The supervised baseline shows low performance, which proves the merit of the learned policies. Overall MCTS performance is comparable to the DQN one. This is mainly due to the learned policies being somewhat local and focusing on object boundaries, not necessitating a large multi-step look-ahead, as we will show in the following.

In \figref{fig:success_cases}, we report success cases of the RL algorithms. Smoothness modeled by our energy fixed the bottle segmentation in the first image. Furthermore, our model detects missing parts of the table in the second image in the first row and a car in the image in the second row, that were missed by the unaries. Also, we show that we fix a mis-labeling of a truck as a car in the image in the third row. 

\begin{figure*}[!h]
\centering
\includegraphics[width=0.8\linewidth]{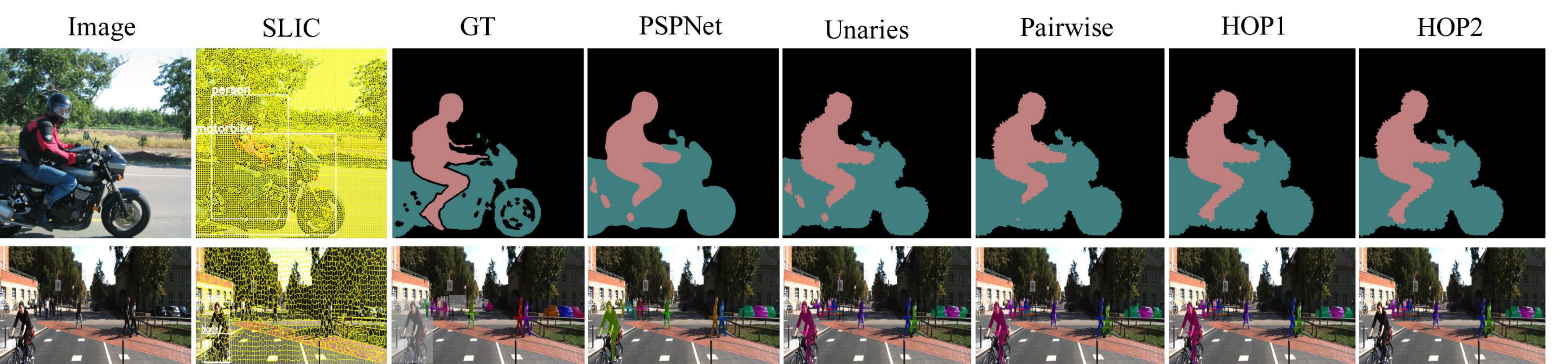}
\vspace{-0.1cm}
\caption{Output of our  method for different potentials.  }
\vspace{-0.5cm}
\label{fig:hoc2}
\end{figure*}

\noindent\textbf{Flexibility of Potentials:} 
In \figref{fig:hoc2}, we show examples of improved segmentations when using the pairwise, HOP1 and HOP2 potentials respectively. The motorcycle driver segmentation improved incrementally with every potential (first image) and the cyclist is not detected anymore as a pedestrian (second image). 

\noindent\textbf{Generalization and scalability:} 
The graph embedding network enables training and testing on graphs with different number of nodes, since the same parameters are used. We investigate how models trained on graphs with few nodes perform on larger graphs. As shown in \tabref{tab:Generalization}, compelling accuracy and IoU values for generalization to graphs with up to 500, 1000, 2000, and 10000 nodes are observed when using a policy trained on graphs of 250 nodes for Pascal VOC, and to graphs with up to 5000 and 10000 nodes when using a policy trained on 2000 nodes for MOTS. Here, we consider the energy consisting of the combined potentials (unary, pairwise, HOP1 and HOP2). Note that we outperform PSPNet at the pixel level for Pascal VOC.
\input{tab_generalization}

\noindent\textbf{Runtime efficiency:} 
In \tabref{tab:Run_Time}, we show the inference runtime for respectively the baselines, DQN and MCTS. The runtime scales \textit{linearly} with the number of nodes and does not even depend on the potential type/order in case of DQN, as inference is reduced to a forward pass of the policy network at every iteration (\figref{fig:policynet}). DQN is faster than all the solvers apart from $\alpha$-exp. However, performance-wise, $\alpha$-expansion has  worse results (\tabref{tab:Unary_Pairwise_Potential}). 
MCTS is slower as it performs multiple simulations per node and requires computation of the reward at every step. 

\input{tab_run_time}

\begin{figure}[t]
\centering
\includegraphics[width=\linewidth]{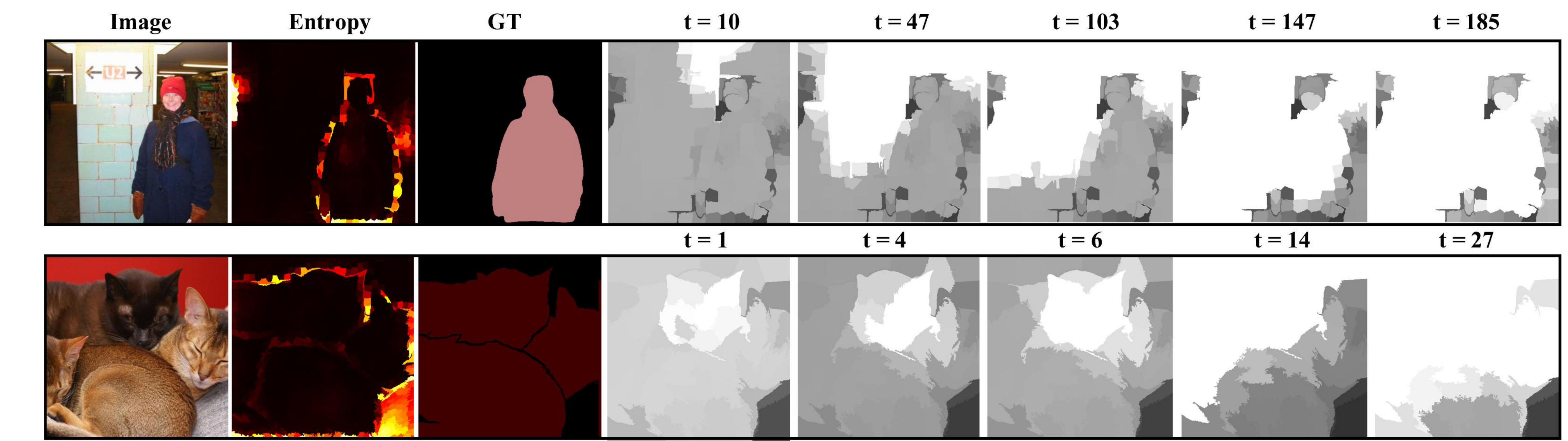}
\vspace{-0.7cm}
\caption{Visualization of the learned policy.}
\label{fig:smoothness1}
\vspace{-0.5cm}
\end{figure}

\noindent\textbf{Learned Policies:} 
In \figref{fig:smoothness1}, we show the probability map across consecutive time steps. The selected nodes are colored in white. The darker the superpixel, the smaller the probability of selecting it next. We found that the heuristic  learns a notion of smoothness, choosing nodes that are in close proximity and of the same label as the selected ones. Also, the policy learns to start labeling the nodes with low unary distribution entropy first, then decides on the ones with  higher entropy.

\begin{figure}[t]
\centering
\includegraphics[width=7cm]{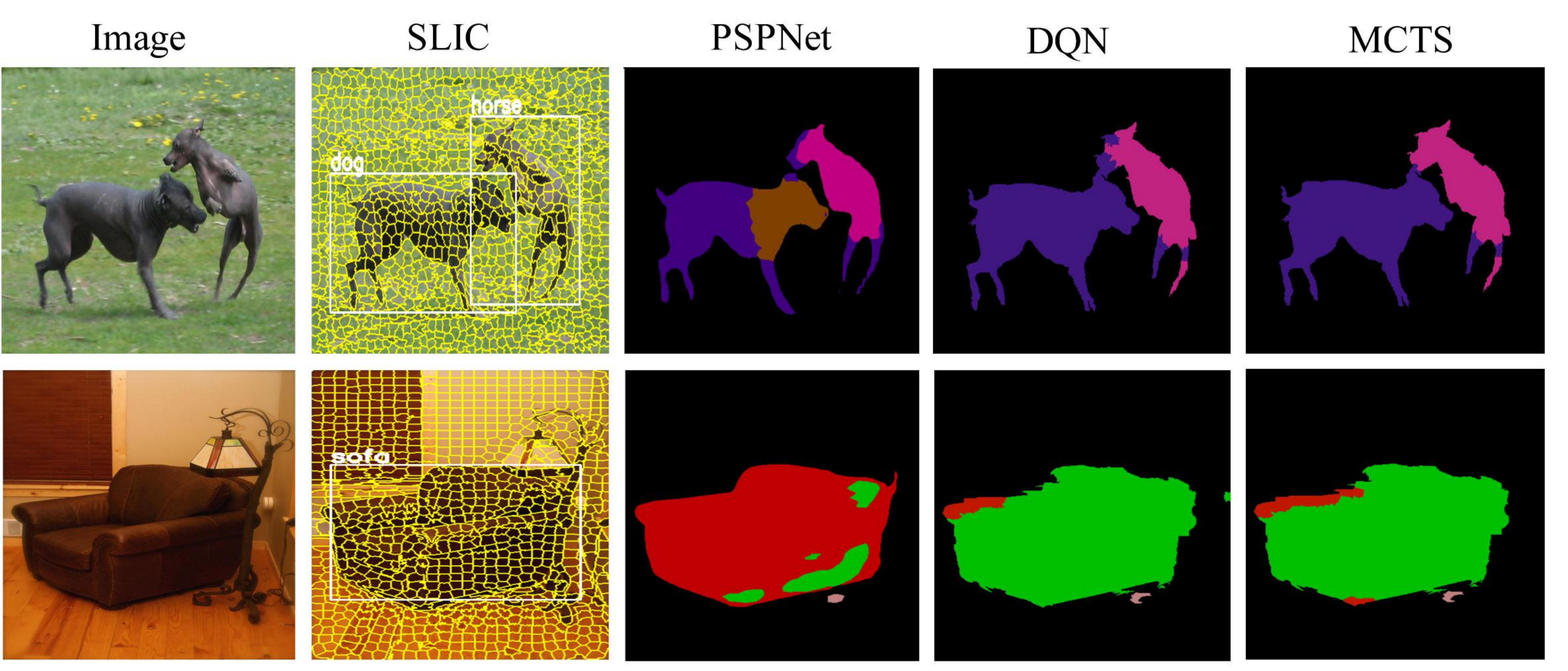}
\vspace{-0.2cm}
\caption{Failure cases.}
\vspace{-0.7cm}
\label{fig:failure_cases}
\end{figure}

\noindent\textbf{Limitations:} Our method is based on super-pixels, hence datasets with small objects require a large number of nodes and a longer run-time (MOTS \vs Pascal VOC). Also, our method is sensitive to bounding box class errors, as illustrated in \figref{fig:failure_cases} (first example), and to the parameters calibration of the energy function, as shown in the second example of the same figure. We plan to address the latter concern in future work via end-to-end training. Furthermore, little is know about deep reinforcement learning convergence. Nevertheless, it has been successfully applied to solve combinatorial programs by leveraging the structure in the data. We show that in our case as well, it converges to reasonable policies. 

%% file: tab_generalization.tex
\begin{table}[t]
\centering
{\tiny
\centering
\caption{\footnotesize{Generalization of the learned policy. }}
\vspace{-0.3cm}
\label{tab:Generalization}
\setlength\tabcolsep{3.5pt}

\begin{tabular}{c|c|c|cc|cc|cc|cc} \toprule
& &  \textbf{PSPNet}  &   \multicolumn{2}{c|}{ \textbf{500}} &  \multicolumn{2}{c|}{ \textbf{1000}}  &   \multicolumn{2}{c|}{ \textbf{2000}} &  \multicolumn{2}{c}{ \textbf{10000}}   \\ \cmidrule(lr){4-5} \cmidrule(lr){6-7} \cmidrule(lr){8-9}\cmidrule(lr){10-11}
\multirow{-2}{*}{\rotatebox[origin=c]{90}{ \textbf{Pascal VOC}}}&&    &    \textbf{ DQN }   &    \textbf{MCTS}  &  \textbf{DQN}    &    \textbf{MCTS}  &  \textbf{DQN}    &    \textbf{MCTS}   &  \textbf{DQN}    &    \textbf{MCTS}      \\  \cmidrule(lr){2-11}
&  \textbf{IoU (sp)} &$-$ & \textbf{88.74}     &   88.73   &   87.58   & \textbf{87.61}  & 86.36  & \textbf{86.39}  & 84.66 & \textbf{84.67}    \\
&  \textbf{IoU (p)}  & 82.61 & \textbf{83.06} &  83.01    &    83.71   & \textbf{83.73}  &  83.74  & \textbf{83.78}  &  83.80 &   \textbf{83.82} \\ 

\toprule
 &&  \textbf{TrackR-CNN}  &     \multicolumn{4}{c|}{ \textbf{5000}} &  \multicolumn{4}{c}{ \textbf{10000}}   \\ \cmidrule(lr){4-7} \cmidrule(lr){8-11}
&&   &    \multicolumn{2}{c}{ \textbf{DQN}}    &    \multicolumn{2}{c|}{ \textbf{MCTS}}  &  \multicolumn{2}{c}{ \textbf{DQN}}    &    \multicolumn{2}{c}{ \textbf{MCTS}}       \\  \cmidrule(lr){2-11}
\multirow{-1.6}{*}{\rotatebox[origin=c]{90}{ \textbf{MOTS} }}&  \textbf{IoU (sp)} &$-$ &  \multicolumn{2}{c}{\textbf{79.81}}   &   \multicolumn{2}{c|}{79.80}   &  \multicolumn{2}{c}{\textbf{76.73}}    &   \multicolumn{2}{c}{76.69} \\
&  \textbf{IoU (p)}  & 84.98 &  \multicolumn{2}{c}{\textbf{83.49}}   &     \multicolumn{2}{c}{83.46}    &  \multicolumn{2}{|c}{\textbf{84.69}}     &   \multicolumn{2}{c}{84.63}    \\ 
\bottomrule
\end{tabular}}
\vspace{-0.3cm}
\end{table}

%% file: tab_run_time.tex
\begin{table}[t]
\centering
{\tiny
\centering
\caption{\footnotesize{Run-time during inference in seconds for Pascal VOC dataset. }}
\vspace{-0.3cm}
\label{tab:Run_Time}
\setlength\tabcolsep{1pt}
\begin{tabular}{c|ccccccc|ccccccc|cc} \toprule
\multirow{-0.1}{*}{\rotatebox[origin=c]{90}{ \textbf{Nodes}}}  &        \multicolumn{7}{c|}{\textbf{U+P}}  &     \multicolumn{7}{c|}{\textbf{U+P+HOP1 }} &  \multicolumn{2}{c}{\textbf{U+P+HOP1}}  \\
 &        \multicolumn{7}{c|}{}  &     \multicolumn{7}{c|}{} &  \multicolumn{2}{c}{\textbf{+HOP2} } \\
\cmidrule(lr){2-8}  \cmidrule(lr){9-15} \cmidrule(lr){16-17} 
 &    \textbf{BP}      & \textbf{TBP}  & \textbf{DD} &\textbf{ L-Flip} & \textbf{$\alpha$-Exp}   &  \textbf{DQN}     & \textbf{MCTS} &      \textbf{BP} & \textbf{TBP}  & \textbf{DD} & \textbf{L-Flip}    & \textbf{$\alpha$-Exp} & \textbf{DQN}    &   \textbf{MCTS}    & \textbf{DQN}    &   \textbf{MCTS}        \\  \midrule
 
\textbf{50}  & 0.14 & 0.52  &  0.28 & 0.12 &\textbf{ 0.01}  & 0.04 &  0.20 &   0.15   & 0.62 &  0.31  &  0.187  & \textbf{0.01} & 0.04 & 0.23 & \textbf{0.04} &   0.24  \\  
\textbf{250} & 1.56 & 2.13 & 1.26  & 0.53 &  \textbf{0.04} & 0.22  & 2.22 &  1.70 & 2.77 &1.65  & 0.59 & \textbf{0.07 } & 0.22  & 2.89 & \textbf{0.22} &   3.01 \\ 
\textbf{500}  & 3.26 &  4.76    & 2.82 & 1.07 & \textbf{0.12} & 0.52 &  7.27  &  3.37 & 5.37  &  3.70 & 0.97 & \textbf{0.22} & 0.53 & 9.17  & \textbf{0.52}  &   9.69   \\
\textbf{1000}  &  6.63 &  9.65&  6.84 & 1.80 & \textbf{0.30}  & 0.78 & 18.5  &   7.22  &10.4    & 7.47 & 2.25  & \textbf{0.36} & 0.78  & 21.6 &  \textbf{0.78} &   22.8  \\ 
\textbf{2000}  &  12.3  &  19.9   &  14.8   &  3.57  & \textbf{0.70} & 1.70   &    38.3  &   12.7  & 23.9  & 15.1    &4.47 & \textbf{0.72} & 1.70 & 43.2 & \textbf{1.72}      &  46.2      \\  
\textbf{10000}  &  72.8 & 130.9  & 143.7 & 22.6 &  \textbf{4.81}   &  8.23  & 202.1  &  88.7 & 140.1  & 106.9 & 23.5    & \textbf{4.72} & 8.25    &  209.7      & \textbf{8.20}    & 210.3    \\
\bottomrule
\end{tabular}}
\vspace{-0.4cm}
\end{table}

%% file: conc.tex
\section{Conclusion}
\label{sec:conc}
\vspace{-0.2cm}
We study how to solve higher order CRF inference for semantic segmentation with reinforcement learning. The approach is able to deal with potentials that are too expensive to optimize using conventional techniques and outperforms traditional approaches while being more efficient. 
Hence, the proposed approach offers more flexibility for energy functions while scaling linearly with the number of nodes and the potential order. To answer our question: can we learn heuristics for graphical model inference? 
We think we can but we also want to note that a lot of manual work is required to find suitable features and graph structures. For this reason we think more research is needed to truly automate learning of heuristics for graphical model inference. We hope the research community will join us in this quest. 

\noindent\textbf{Acknowledgements:} 
This work is supported in part by NSF under Grant No.\ 1718221 and MRI \#1725729, UIUC, Samsung, 3M, and Cisco Systems Inc.\ (Gift Award CG 1377144). 
We thank Cisco for access to the Arcetri cluster and Iou-Jen Liu for initial discussions.


%% file: supplementary.tex
\centerline{\LARGE{\textbf{Supplementary Material }}}
\bigbreak

Recall, given an image $x$, we are interested in predicting the semantic segmentation $y = (y_1, \ldots, y_N) \in \cY$ by solving the inference task defined by a Conditional Random Field (CRF) with nodes corresponding to superpixels. Hereby, $N$ denotes the total number of superpixels. The semantic segmentation of a superpixel $i\in\{1, \ldots, N\}$ is referred to via $y_i\in\cL = \{1, \ldots, |\cL|\}$, which can be assigned one out of $|\cL|$ possible discrete labels from the set of possible labels $\cL$. We formulate the inference task as a Markov Decision Process that we study using two reinforcement learning algorithms: DQN and MCTS.
Specifically, an agent operates in $t\in\{1, \ldots, N\}$ time-steps according to a policy $\pi(a_t | s_t)$ which encodes a probability distribution over actions $a_t\in\cA_t$ given the current state $s_t$. The current state subsumes the indices of all currently labeled variables $I_t\subseteq\{1, \ldots, N\}$ as well as their labels $y_{I_t} = (y_i)_{i\in I_t}$, \ie, $s_t \in \{ (I_t, y_{I_t}) : I_t \subseteq \{1, \ldots, N\}, y_{I_t} \in \cL^{|I_t|}\}$. The policy selects one superpixel and its corresponding label at every time-step.\\

In this supplementary material we present:
\begin{enumerate}
\item \textbf{Appendix A}: Further training and implementation details
\item \textbf{Appendix B}: Further details on the policy network 
\begin{itemize}
\item \textbf{B 1}: DQN
\item \textbf{B 2}: MCTS
\end{itemize}
\item \textbf{Appendix C}: Additional Results 
\begin{itemize}
\item \textbf{C 1}: Comparison of the reward schemes
\item \textbf{C 2}: Visualization of the learned embeddings
\item \textbf{C 3}: Learned policies
\item \textbf{C 4}: Qualitative results 
\end{itemize}
\end{enumerate}

\input{RL_Seg_CVPR_Supp/appendixA}

\input{RL_Seg_CVPR_Supp/appendixB}
\newpage
\input{RL_Seg_CVPR_Supp/appendixC}

%% file: RL_Seg_CVPR_Supp/appendixA.tex

\section*{A: Further Training and Implementation Details}
PSPNet, TrackR-CNN and and the hypercolumns from VGGNet are not fine-tuned. Only the graph policy net is trained. For MOTS, we apply our model to every frames of the video. For MCTS, we set the number of simulations during exploration to 50 and the simulation depth to 4. At test time, we run 20 simulations with a depth of 4. The models are trained for 10 epochs, equivalent to around $375,000$ training iterations for Pascal VOC and $188,000$ for MOTS. The parameters of the energy function, $\alpha_{p}$, $\beta_{p}$, $w_{b}$, $c_{b}$, $\lambda_{b}$ and $C$, are obtained via a grid search on a subset of 500 nodes from the training data. The number of iterations $K$ of the graph neural network is set to 3. The dimension $F$ of the node features $b_{i}$ equals $85$ for Pascal VOC and $30$ for MOTS, consisting of the unary distribution, the unary distribution entropy, and features of the bounding box.  The bounding box features are the confidence and label of the bounding box, its unary composition at the pixel level, percentage of overlap with other bounding boxes and their associated labels and confidence. The embedding dimension $p$ is $32$ for Pascal VOC and $16$ for MOTS. As an optimizer, we use Adam with a learning rate of $0.001$.

%% file: RL_Seg_CVPR_Supp/appendixB.tex
\section*{B: Further details on the policy network (DQN, MCTS) }
\subsection*{B1: DQN}
Both for DQN and MCTS, three different sets of actions are encouraged at training iteration $t$: 
\begin{itemize}
\item $\mathcal{M}_{1}^{(t)}$: Selecting nodes adjacent to the already chosen ones in the graph, at iteration $t$. Otherwise, the reward will only be based on the unary terms as the pairwise term is only evaluated if the neighbors are labeled ($t = 2$ in Tab. 1 in the main paper).


We assign a score $M_{1}(s_{t},a_{t})$ to every available action $a_{t}=(i_{t},y_{i_t})\in \mathcal{A}_{t}$ to encourage the exploration of the set $\mathcal{M}_{1}^{(t)}$:

\be
M_{1}(s_{t},a_{t})= \frac{| \{j: (j \notin I_{t}) \text{ and } (i_{t}, j) \in \cal{E} \}  |}{| \{j: (i_{t}, j) \in \cal{E}\} |}. 
\ee

\item $\mathcal{M}_{2}^{(t)}$: Selecting nodes with the lowest unary distribution entropy, at iteration $t$. A low entropy indicates a high confidence of the unary deep net. Hence, the labels of the corresponding nodes are more likely to be correct and would provide useful information to neighbors with higher entropy in the upcoming iterations. 
We assign a score $M_{2}(s_{t},a_{t})$ to every available action $a_{t}=(i_{t},y_{i_t})\in \mathcal{A}_{t}$ to encourage the exploration of the set $\mathcal{M}_{2}^{(t)}$:

\be
M_{2}(s_{t},a_{t})= \frac{ \exp{(-S_{i_{t}})}  }{ \sum_{   j \in \{1,\dots,N\} \setminus I_{t}  }  \exp{(-S_j)}   }.
\ee
Here, $S_{i_t}$ denotes the entropy of the unary distribution evaluated at node $i_t$.
\item $\mathcal{M}_{3}^{(t)}$: Assigning the same label to nodes forming the same higher order potential at iteration $t$, \ie, 
\be
M_{3}(s_{t},a_{t})=\begin{cases}
      1 & \text{if }  y_{i_t} = \argmax\limits_{k\in\cL}  \sum\limits_{  \{j : j \in I_{t} \text{ and } (i_{t},j) \in \cal{C}\} } \mathbbm{1}_{\{y_j=k\}}     \\
      0 & \text{otherwise}
    \end{cases}. 
\ee

\end{itemize}

For \textbf{DQN}, at train time, the next action $a_{t}$ is selected as follows:
\be
a_{t}^{*}= \begin{cases}
       \argmax\limits_{a_{t}\in \mathcal{A}_{t}}Q(s_{t}, a_{t}; \theta) & \text{with probability }\epsilon\\
       \argmax\limits_{a_{t}\in \mathcal{A}_{t}}M_{1}(s_{t},a_{t})& \text{with probability } (1-\epsilon)/4\\
       \argmax\limits_{a_{t}\in \mathcal{A}_{t}}M_{2}(s_{t},a_{t})& \text{with probability } (1-\epsilon)/4\\
       \argmax\limits_{a_{t}\in \mathcal{A}_{t}}M_{3}(s_{t},a_{t})& \text{with probability } (1-\epsilon)/4\\
      \text{Random}& \text{with probability } (1-\epsilon)/4.
    \end{cases} 
\ee
Here $\epsilon$ is a fixed probability modeling the exploration-exploitation tradeoff. At test time, $a_{t}^{*}= \argmax\limits_{a_{t}\in\mathcal{A}_{t}} Q(s_{t}, a_{t}; \theta)$.



\subsection*{B2: MCTS}
As described in Sec 3.5\ of the main paper, for a given graph $G(V,\mathcal{E}, w)$, MCTS operates by constructing a tree, where every node corresponds to a state $s$ and every edge corresponds to an
action $a$. The root node is initialized to $s_{1}=\emptyset$. Every node stores three statistics: 
1) $N(s)$, the number of times state $s$ has been reached, 
2) $N(a|s)$, the number of times action $a$ has been chosen in node $s$ in all previous simulations, and
3) $\tilde{r}(s,a)$, the averaged reward across all simulations starting at state $s$ and taking action $a$. 
A simulation involves three steps : 1) selection, 2) expansion and 3) value backup. After running $n_\text{sim}$ simulations, an empirical distribution  $\pi^{\text{MCTS}}(a|s)= \frac{N(a|s)}{N(s)}$ is computed for every node. The next action is then chosen according to $\pi^{\text{MCTS}}$. A policy network $\pi_\theta(a|s)$ is trained to match a distribution $\pi^{\text{MCTS}}$ constructed through these simulations. In the following, we provide more details about each of these steps.

\noindent \textbf{Selection} corresponds to choosing the next action given the current node $s_t$, based on four factors : 1) a variant of the probabilistic upper confidence bound (PUCB) given by $U(s_{t}, a_{t}; \theta)= \frac{\tilde{r}(s_t,a_t)}{N(a_t|s_t)} +  \pi_\theta(a_t|s_t) \frac{\sqrt{N(s_t)}}{1 + N(a_t|s_t)}$, 2) $M_1(s_{t},a_{t})$ 3) $M_2(s_{t},a_{t})$ and 4) $M_3(s_{t},a_{t})$ similarly to  DQN in \textbf{Appendix B1}. Formally, 
\be
a_{t}^{*}= \argmax\limits_{a_{t}\in\mathcal{A}_{t}} \begin{cases}
      U(s_{t}, a_{t}; \theta) + M_{1}(s_{t},a_{t})& \text{with probability } \frac{1}{3}\\
      U(s_{t}, a_{t}; \theta) + M_{2}(s_{t},a_{t})& \text{with probability } \frac{1}{3}\\
      U(s_{t}, a_{t}; \theta) + M_{3}(s_{t},a_{t})& \text{with probability } \frac{1}{3}
    \end{cases}. 
\label{eq:exploref}
\ee

\noindent \textbf{Expansion} consists of constructing a child node for every possible action from the parent node $s_t$. The possible actions include the nodes which have not been labeled. The child nodes' cumulative rewards and counts are initialized to 0. Note that selection and expansion are limited to a depth $d_\text{sim}$ starting from the root node in a simulation. 

\noindent \textbf{Value backup} refers to back-propagating the reward from the current node on the path to the root of the sub-tree. The visit counts of all the nodes in the path are incremented as well.


\noindent \noindent \textbf{Final Labeling}: Once $n_\text{sim}$ simulations are completed, we compute $\pi^{\text{MCTS}}$ for every node. The next action $a_t$ from the root node is decided according to $\pi^{\text{MCTS}}(a_t|s_t)$ : $a_t \sim \pi^{\text{MCTS}}(a_t|s_t)$ at train time and $a_t = \argmax_{a_t\in\mathcal{A}_{t}} \pi^{\text{MCTS}}(a_t|s_t) $ at inference. The next node becomes the root of the sub-tree. The experience ($s_t$, $\pi^{\text{MCTS}}$) is stored in the replay buffer. The whole process is repeated until all $N$ nodes in the graph $G$ are labeled.  We summarize the MCTS training algorithm below in \algref{alg:ours_detailed}. Note that we run 10 episodes per graph during training, but for simplicity we present the training for a single episode per graph.



\begin{algorithm}[!h]
\SetKwInOut{Input}{input}
\SetKwInOut{Output}{output}
\Input{Head node: $s_1$, $n_\text{sim}$: number of simulations, $d_\text{sim}$ : depth of simulations}
\Output{A labeling $y\in\mathcal{Y}$ for all the nodes $V$}
\BlankLine
\tcp{Looping over the graphs from the dataset}
\For{ \textbf{all} $G(V,\mathcal{E}, w)$}{
\tcp{Initialization}
$s_1 = \emptyset$ \\
$\tilde{r}(s_{i},a)= 0 $, $ \forall s_i, i \in \{1,\dots, N\}, \forall a$\\
\tcp{Looping over graph nodes $V$}
\For{$t = 1$ \KwTo $N$}{\label{Level_Move}
\tcp{Running simulations}
\For{$n = 1$ \KwTo $n_\text{sim}$}{\label{Simulations}
\tcp{Create and expand a sub-tree}
\For{$j = t$ \KwTo $t+d_\text{sim}$}{\label{Levels}
Select $a_j$ according to \equref{eq:exploref} and advance temporary state in sub-tree\\
}
Backup rewards along the visited nodes in the simulations\\
Update node visit counts\\
}
Compute tree policy $\pi^{\text{MCTS}}$ with visit counts\\
Select the next action $a_t \sim \pi^{\text{MCTS}}(a_t|s_t)$\\
Update the root node $s_{t+1}$ $\leftarrow$ $s_{t} \oplus a_t$\\
Store $(s_t, \pi^{\text{MCTS}})$ in Replay Buffer\\

}

Sample $M$ examples from Replay Buffer to update neural network parameters using \textbf{Eq.~(4)} in the main paper\\

}\caption{Monte Carlo Tree Search Training}
\label{alg:ours_detailed}
\end{algorithm}

%% file: RL_Seg_CVPR_Supp/appendixC.tex
\section*{C: Additional  Results}
\subsection*{C1: Comparing Reward Schemes}

\begin{figure}[!h]
\centering
\includegraphics[width=10cm]{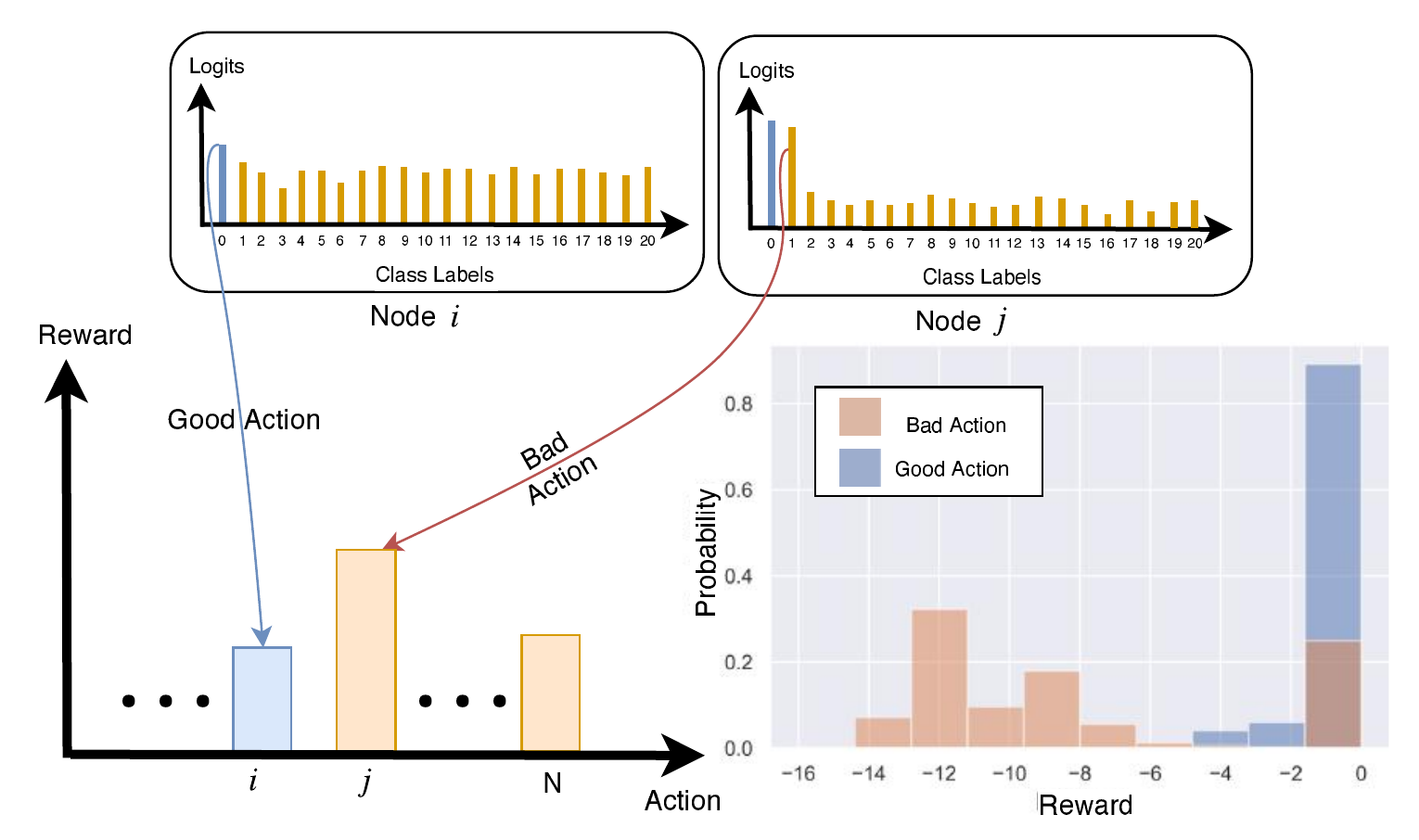}
\vspace{-0.2cm}
\caption{ Explanation of the low performance of the first reward scheme ($r_{t} = - E_{t} + E_{t-1}$). }
\label{fig:reward_dist}
\end{figure}
 
In Tab.~2 of the main paper, we observe that the second reward scheme ($r_{t}=\pm 1$) generally outperforms the first one ($r_{t}=-E_{t}+E_{t-1}$). This is due to the fact that, the rewards for wrong actions, in this scheme, can be higher than the ones for good actions. Specifically, in \figref{fig:reward_dist} we plot the distribution of the rewards of good actions (in blue) and the one of wrong actions (in orange) for $50,000$ randomly chosen actions from the replay memory. To better illustrate the cause, we consider the unary energy  case and visualize the class distribution of two nodes $i$ and $j$. If we label node $i$ to be of class 0 (good action), and node $j$ to be of class 1 (wrong action), the resulting rewards are $f_i(0)$ for node $i$ and $f_j(1)$ for node $j$. Note that $f_i(0)<f_j(1)$, since the distribution of the labels in case of node $i$ is almost uniform, whereas the mass for node $j$ is put on the first two labels.

\subsection*{C2: Visualization of the learned embeddings} 
\begin{figure}[!h]
\centering
\includegraphics[width=8cm]{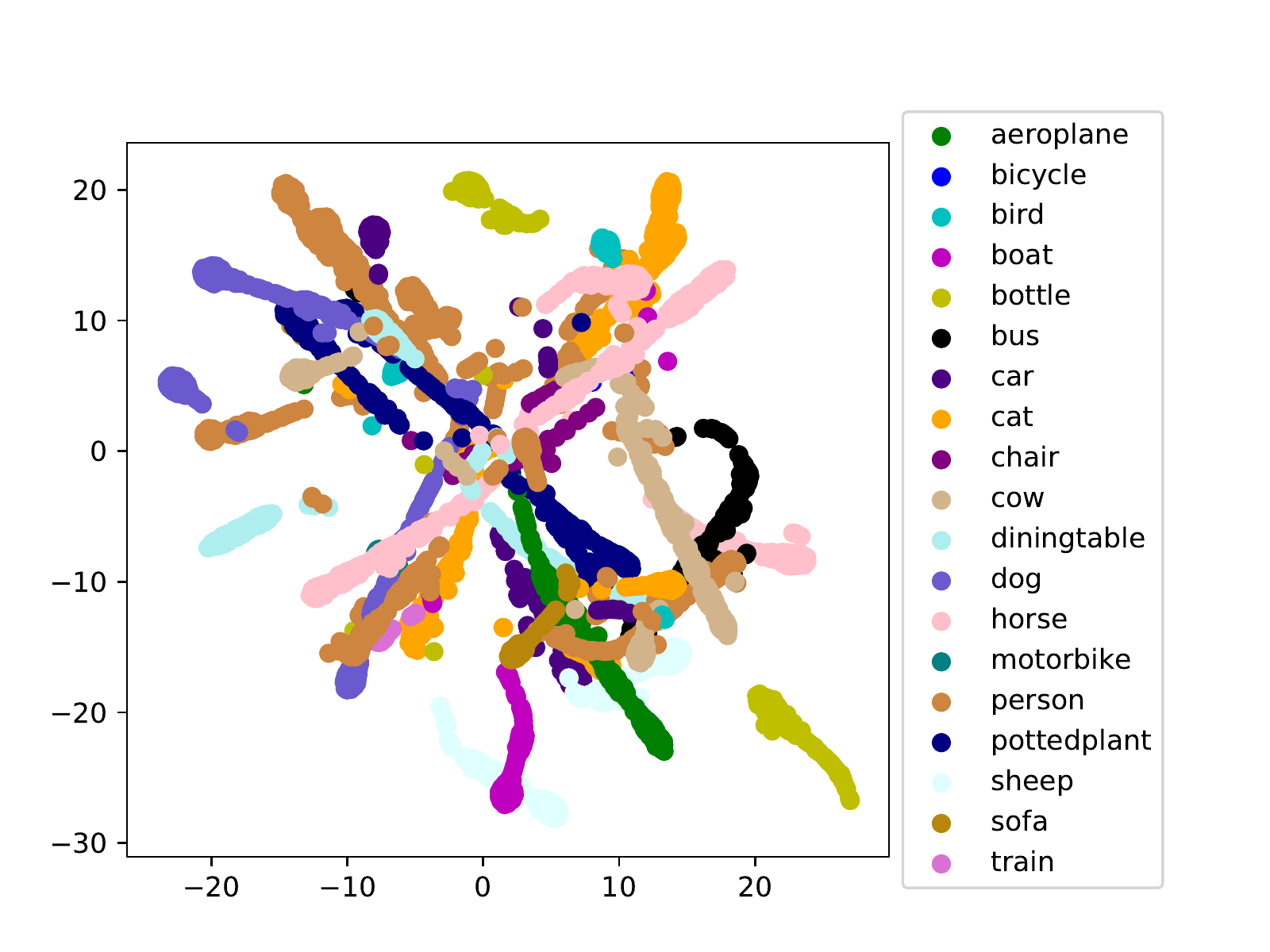}
\caption{Visualization of the learned embeddings.}
\label{fig:embeddings}
\end{figure}

In Tab.~2 in the main paper, we observe that our model can produce better segmentations than the ones obtained by just optimizing energies. Guided by the reward and due to network regularization, the policy net captures contextualized embeddings of classes beyond energy minimization. Intuitively, a well calibrated energy function yields rewards that are well correlated with F1-scores for segmentation. When TSNE-projecting the policy nets node embeddings for Pascal VOC data into a 2D space, we observe that they cluster in 21 groups, as illustrated in \figref{fig:embeddings}.

\clearpage

\subsection*{C3: Learned Policies} 
\begin{figure}[!h]
\centering
\includegraphics[width=15cm]{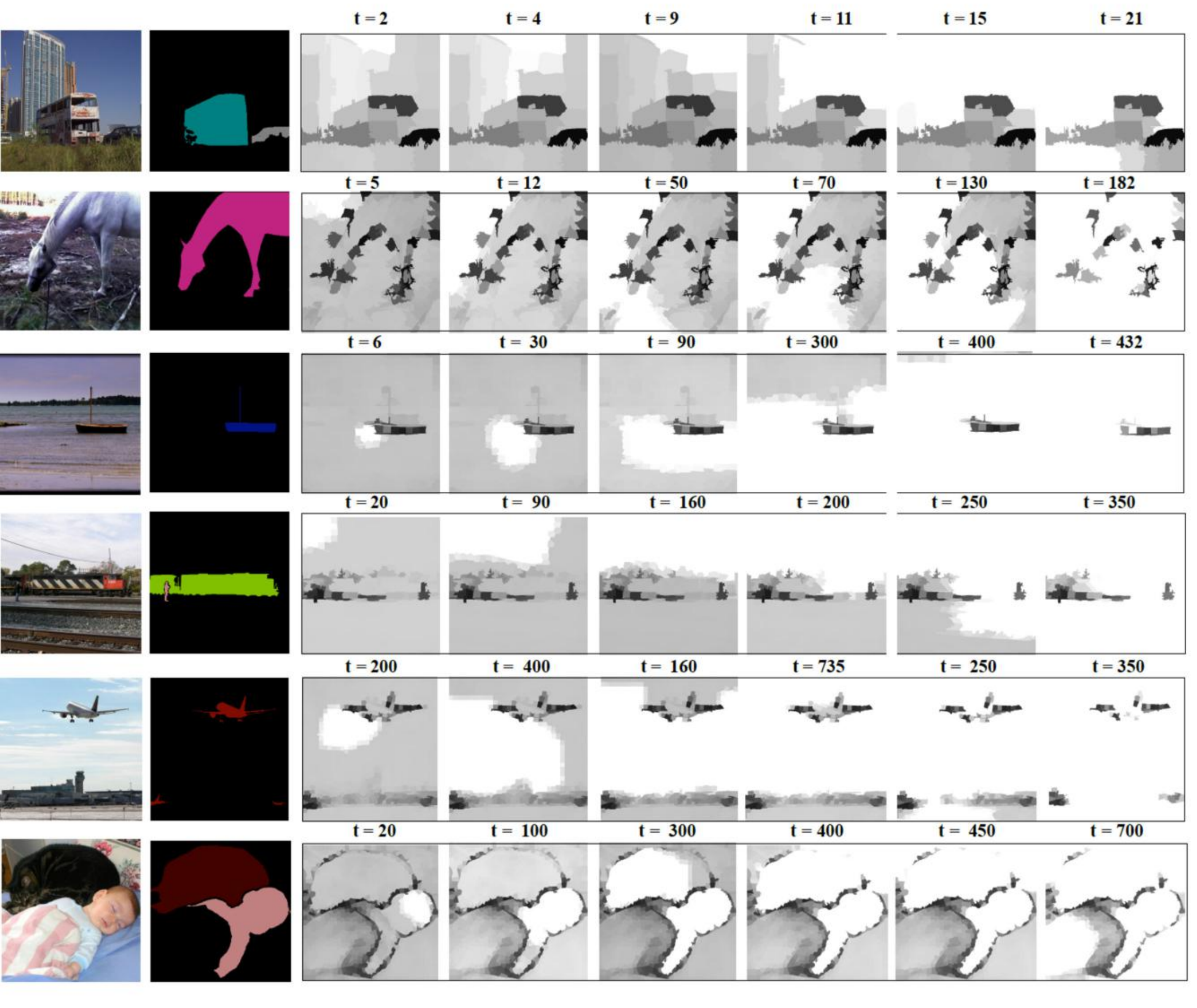}
\caption{Visualization of our learned policy. }
\label{fig:policynet}
\end{figure}

In \figref{fig:policynet}, we visualize the learned greedy policy. Specifically, we show the probability map across consecutive time steps. The probability maps are obtained by first computing an $N$ dimensional score vector $\phi((i,\cdot)|s_{t}) = \sum_{y\in\cL} \pi((i,y)|s_{t}) $ $\forall i\in\{1, \ldots, N\}$ by summing over all the label scores per node and then normalizing $\phi((i,\cdot)|s_{t})$ to a probability distribution over the non selected superpixels $i\in\{1, \ldots, N\}\setminus I_t$.
The selected nodes are colored in white. The darker the superpixel, the smaller the probability of selecting it next. We found that the heuristic  learns a notion of smoothness, selecting nodes that are in close proximity and of the same label as the selected ones. Also, the policy learns to start labeling the nodes with low unary distribution entropy, then decides on the ones with  higher entropy.

\subsection*{C4: Additional qualitative results} 
\begin{figure}[!h]
\centering
\includegraphics[width=17cm]{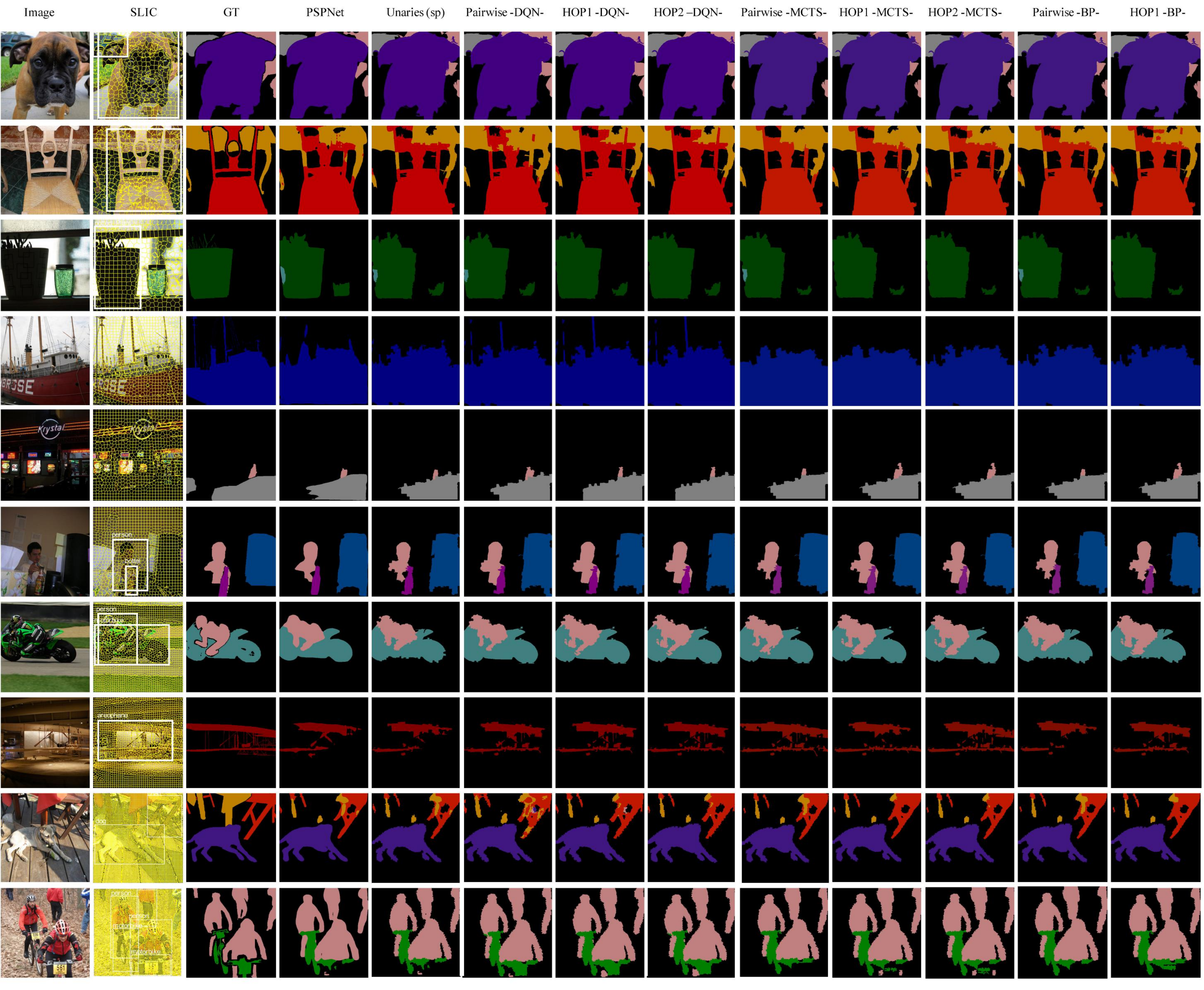}
\caption{Output of our method for different potentials for Pascal VOC. }
\label{fig:potentials}
\end{figure}

\begin{figure}[!h]
\centering
\includegraphics[width=17cm]{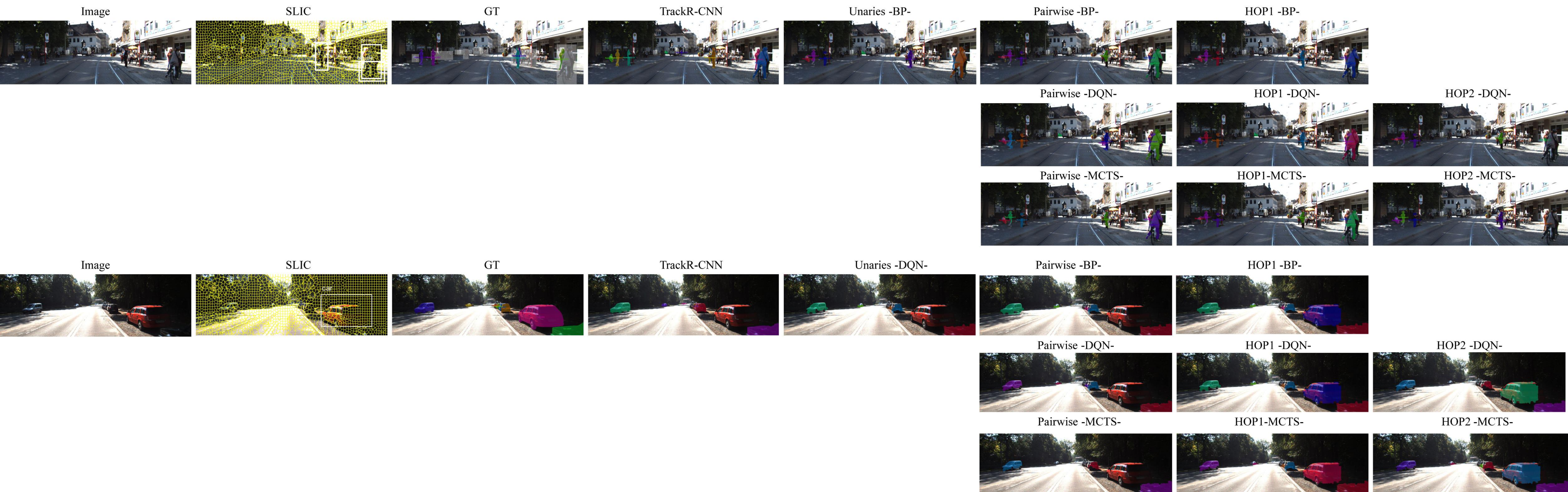}
\caption{Output of our method for different potentials for MOTS. }
\label{fig:potentials_mots}
\end{figure}

In the following, we present additional qualitative results.  In \figref{fig:potentials} and \figref{fig:potentials_mots}, we present the segmentation results of our policy on examples from the Pascal VOC and the MOTS datasets respectively. The pairwise potential, together with the superpixel segmentation helped reduce inconsistencies in the unaries obtained from PSPNet/TrackR-CNN across all the examples. HOP1 resulted in better learning the boundaries of the objects. The energy which includes the HOP2 potential provides the best results across all energies as it helped better segment overlapping objects.

\begin{figure}[!h]
\centering
\includegraphics[width=18cm]{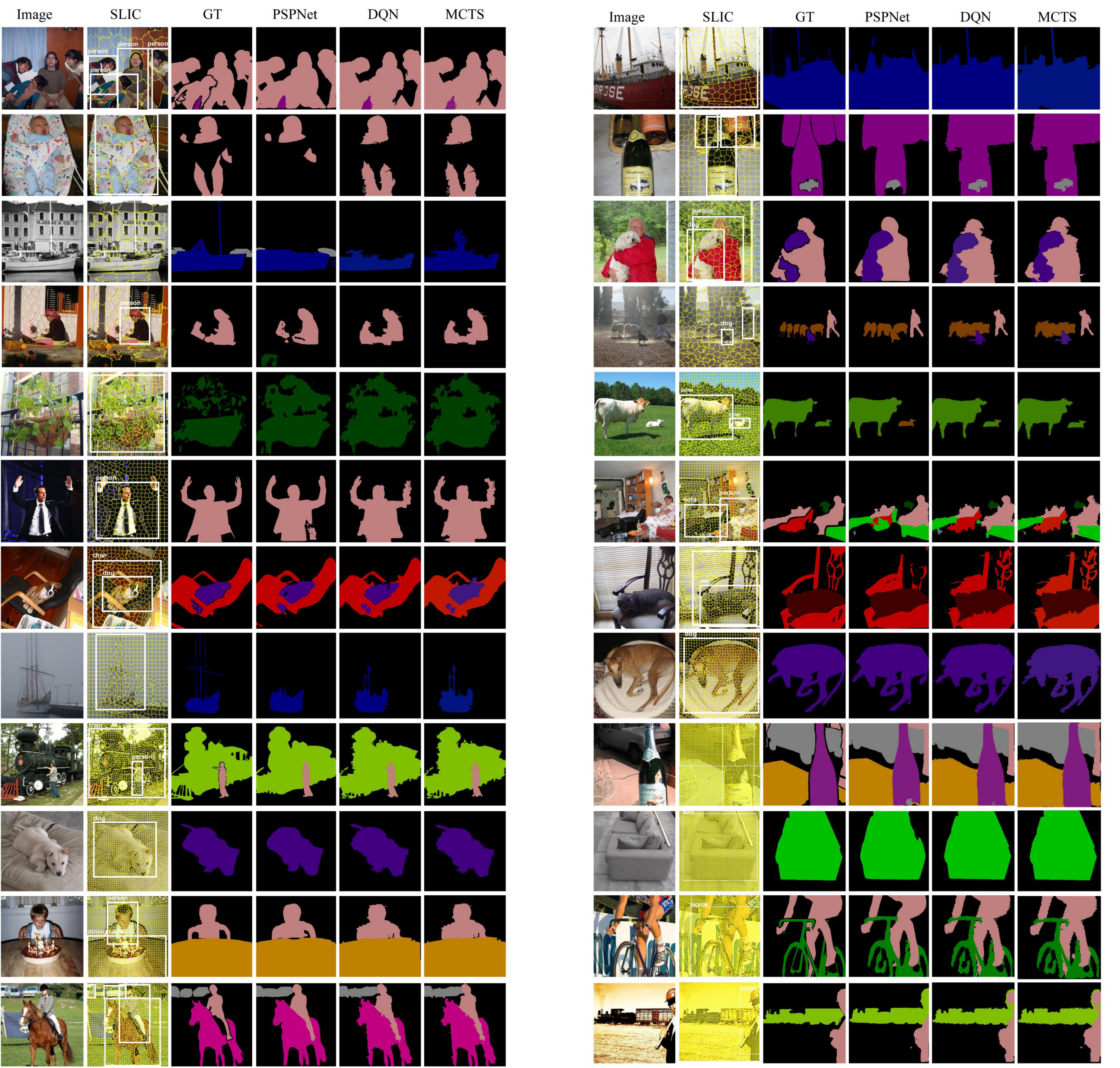}
\caption{Additional success cases on Pascal VOC. }
\label{fig:success_voc}
\end{figure}

\begin{figure}[!h]
\centering
\includegraphics[width=18cm]{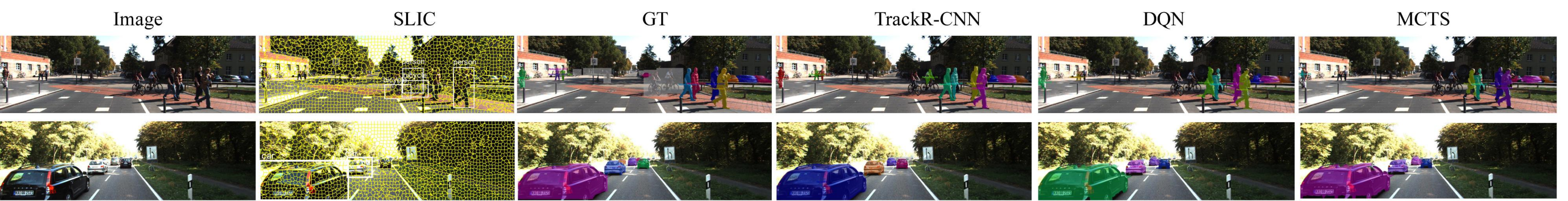}
\caption{Additional success cases on MOTS.}
\label{fig:success_mots}
\end{figure}

\begin{figure}[!h]
\centering
\includegraphics[width=18cm]{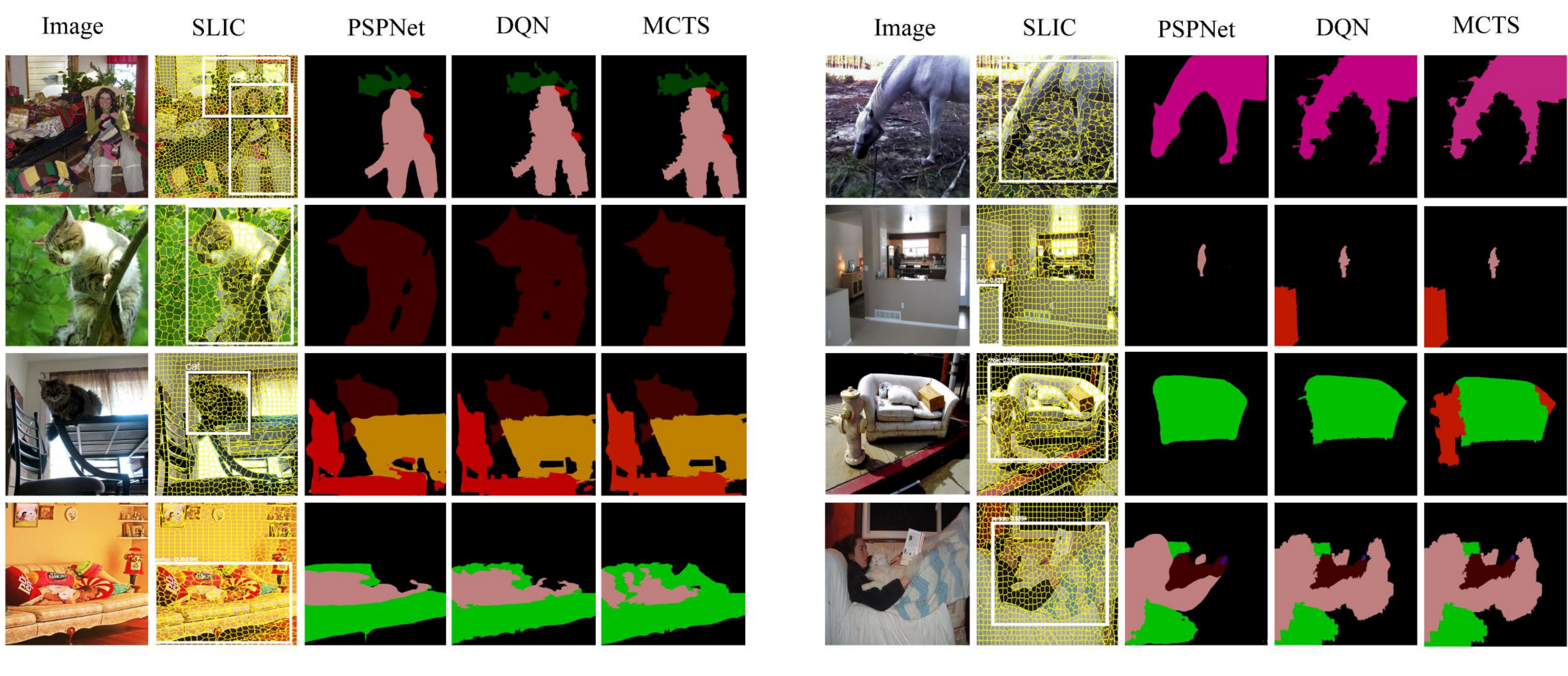}
\caption{ Additional failure cases on Pascal VOC. }
\label{fig:failure_voc}
\end{figure}

\begin{figure}[!h]
\centering
\includegraphics[width=18cm]{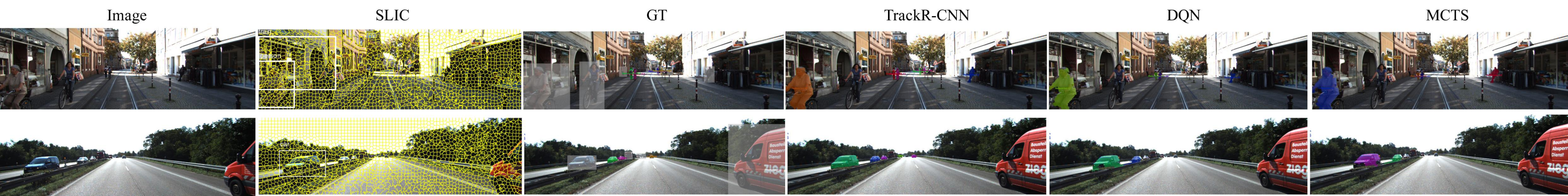}
\caption{ Additional failure cases on MOTS. }
\label{fig:failure_mots}
\end{figure}

We include additional results comparing PSPNet/TrackR-CNN, DQN and MCTS outputs for the energy function with unary, pairwise, HOP1 and HOP2 potentials in \figref{fig:success_voc} and \figref{fig:success_mots}. The policies trained with DQN/MCTS improve over the PSPNet/TrackR-CNN results across almost all our experiments. Additional failure cases are presented in \figref{fig:failure_voc} and \figref{fig:failure_mots}.

